%% file: IDiscMain.tex
\documentclass[10pt,twocolumn,letterpaper]{article}

\usepackage[pagenumbers]{cvpr} % To force page numbers, \eg for an arXiv version

\usepackage{microtype}
\usepackage{graphicx}
\usepackage{amsmath}
\usepackage{amssymb}
\usepackage{mathtools}
\usepackage{booktabs}
\usepackage{multirow}
\usepackage{pifont}
\usepackage{tabularx}
\usepackage{color}
\usepackage{xcolor}
\usepackage{subcaption}
\usepackage{placeins}
\usepackage[accsupp]{axessibility}
\usepackage[pagebackref,breaklinks,colorlinks]{hyperref}

\newcommand{\cmark}{\ding{51}}
\newcommand{\xmark}{\ding{55}}
\newcommand*{\ourmodule}{ID\@\xspace}
\newcommand*{\ourmodulename}{Internal Discretization\@\xspace}
\newcommand*{\ourmodel}{iDisc\@\xspace}

\usepackage[capitalize]{cleveref}
\Crefname{section}{Section}{Sections}
\Crefname{table}{Table}{Tables}
\crefname{section}{Sec.}{Secs.}
\crefname{table}{Tab.}{Tabs.}

\def\cvprPaperID{1920} % *** Enter the CVPR Paper ID here
\def\confName{CVPR}
\def\confYear{2023}

\begin{document}

%%%%%%%%% TITLE - PLEASE UPDATE
\title{iDisc: Internal Discretization for Monocular Depth Estimation}
\author{
Luigi Piccinelli%
\qquad Christos Sakaridis%
\qquad Fisher Yu\vspace{6px}\\%
Computer Vision Lab, ETH Z\"urich
}
\maketitle

\input{IDiscPaper}

\clearpage
%%%%%%%%% REFERENCES
{\small
\bibliographystyle{ieee_fullname}
\bibliography{IDiscMain}
}
Appendix
\clearpage
\appendix
\input{IDiscAppendix}

\end{document}

%% file: IDiscPaper.tex
%%%%%%%%% ABSTRACT

\begin{abstract}
Monocular depth estimation is fundamental for 3D scene understanding and downstream applications. However, even under the supervised setup, it is still challenging and ill-posed due to the lack of full geometric constraints. Although a scene can consist of millions of pixels, there are fewer high-level patterns. We propose \ourmodel to learn those patterns with internal discretized representations. The method implicitly partitions the scene into a set of high-level patterns. 
In particular, our new module, \ourmodulename (\ourmodule), implements a continuous-discrete-continuous bottleneck to learn those concepts without supervision. In contrast to state-of-the-art methods, the proposed model does not enforce any explicit constraints or priors on the depth output. 
The whole network with the \ourmodule module can be trained end-to-end, thanks to the bottleneck module based on attention. Our method sets the new state of the art with significant improvements on NYU-Depth v2 and KITTI, outperforming all published methods on the official KITTI benchmark. \ourmodel can also achieve state-of-the-art results on surface normal estimation. 
Further, we explore the model generalization capability via zero-shot testing. We observe the compelling need to promote diversification in the outdoor scenario. Hence, we introduce splits of two autonomous driving datasets, DDAD and Argoverse. Code is available at \url{http://vis.xyz/pub/idisc}.
\end{abstract}

%%%%%%%%% BODY TEXT
\vspace{-10pt}
\section{Introduction}
\label{sec:intro}

\begin{figure}[ht]
    \centering
    \begin{subfigure}[b]{0.495\columnwidth}
        \centering
        \includegraphics[width=1\linewidth]{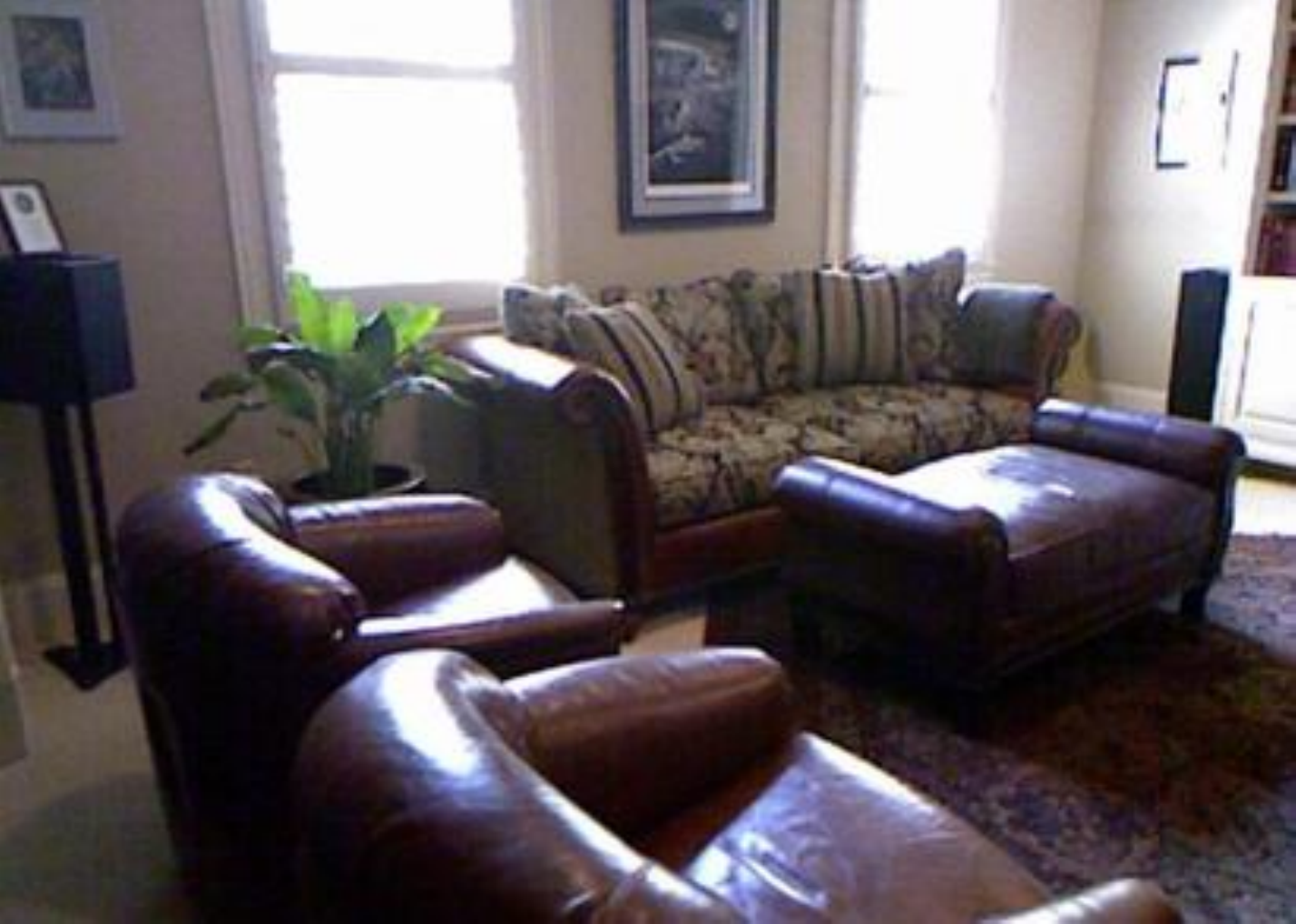}
        \caption{Input image}
    \end{subfigure}
    \begin{subfigure}[b]{0.495\columnwidth}
        \centering
        \includegraphics[width=1\linewidth]{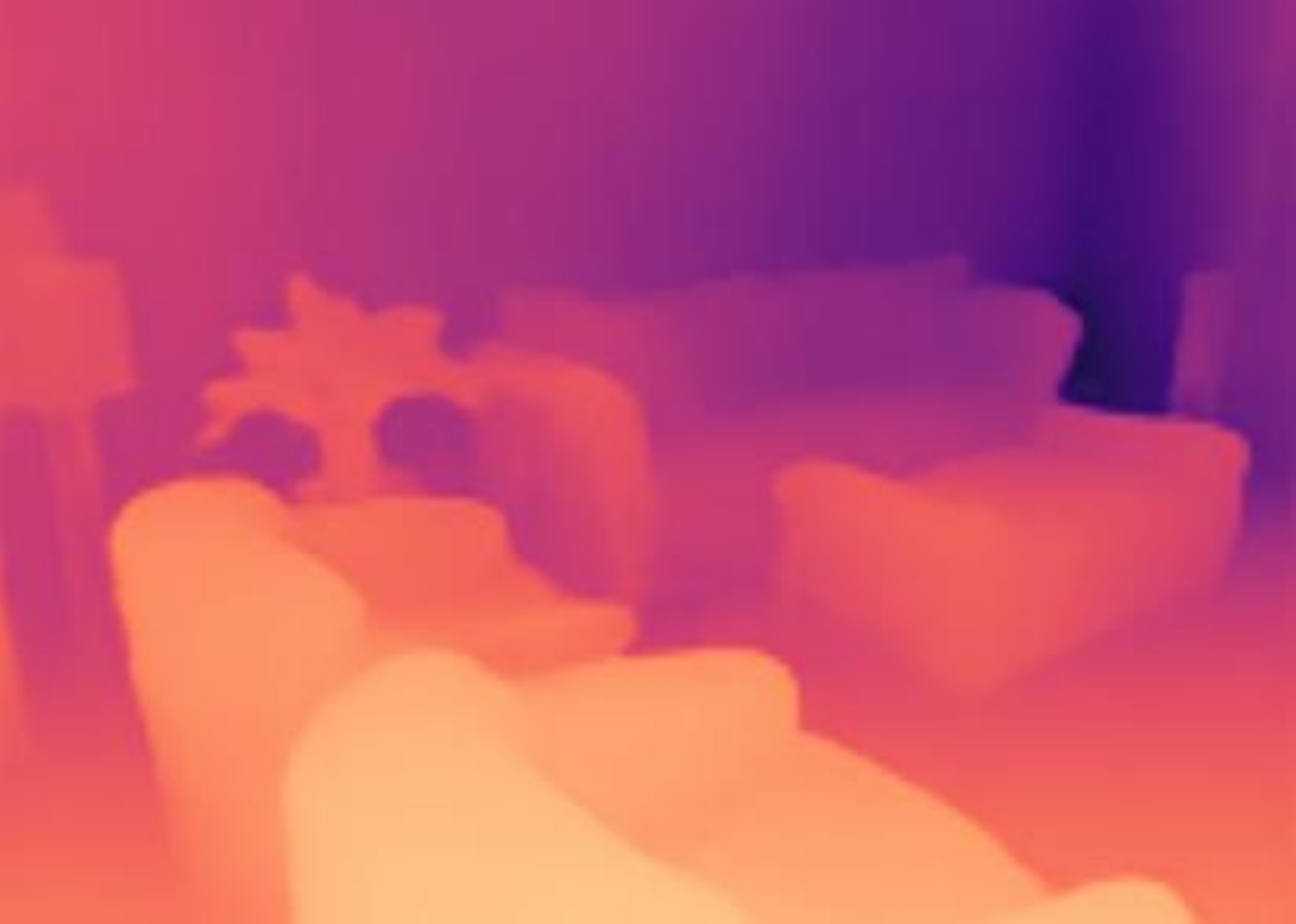}
        \caption{Output depth}
    \end{subfigure}
    \par\smallskip
    \begin{subfigure}[b]{0.495\columnwidth}
        \centering
            \includegraphics[width=1\linewidth]{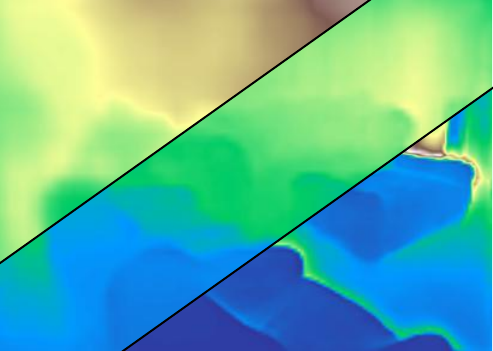}
        \caption{Intermediate representations}
         \label{fig:intermediate_repr}
    \end{subfigure}
    \begin{subfigure}[b]{0.495\columnwidth}
        \centering
        \includegraphics[width=1\linewidth]{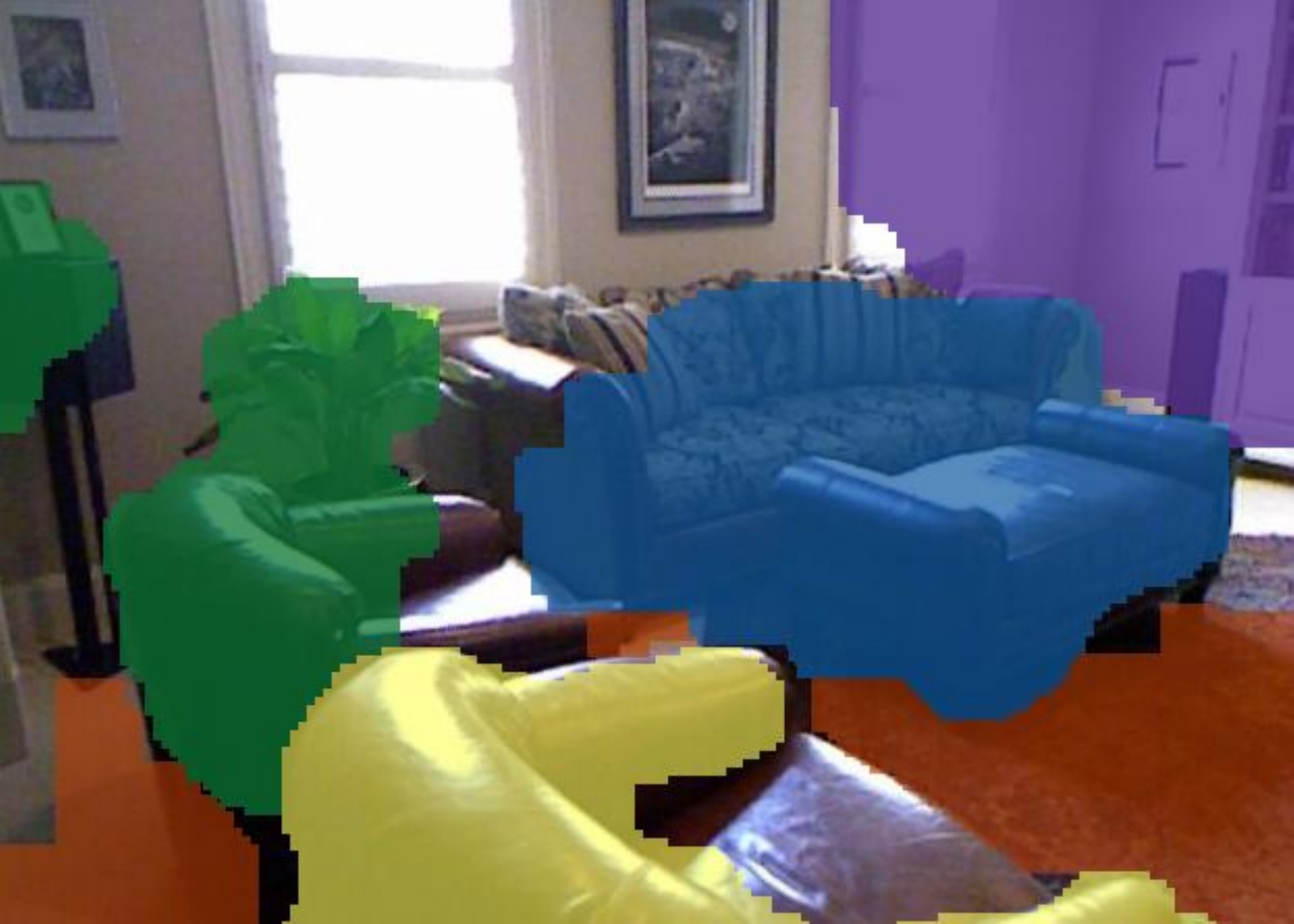}
        \caption{Internal discretization}
        \label{fig:internal_disc}
    \end{subfigure}\\
    \vspace{-5pt}
    \caption{
    We propose \ourmodel which implicitly enforces an internal discretization of the scene via a continuous-discrete-continuous bottleneck. Supervision is applied to the output depth only, \ie, the fused intermediate representations in \subref{fig:intermediate_repr}, while the internal discrete representations are implicitly learned by the model. \subref{fig:internal_disc} displays some actual internal discretization patterns captured from the input, \eg, foreground, object relationships, and 3D planes. Our \ourmodel model is able to predict high-quality depth maps by capturing scene interactions and structure.}
\label{fig:teaser}
\vspace{-15pt}
\end{figure}

Depth estimation is essential in computer vision, especially for understanding geometric relations in a scene. This task consists in predicting the distance between the projection center and the 3D point corresponding to each pixel. Depth estimation finds direct significance in downstream applications such as 3D modeling, robotics, and autonomous cars. Some research~\cite{Zhou2019} shows that depth estimation is a crucial prompt to be leveraged for action reasoning and execution. In particular, we tackle the task of monocular depth estimation (MDE). MDE is an ill-posed problem due to its inherent scale ambiguity: the same 2D input image can correspond to an infinite number of 3D scenes.

% State-of-the-Art (SotA) methods typically involve convolutional networks that enable smooth and fine-grained depth prediction~\cite{Eigen2014, Fu2018, Lee2019}. Since the advent of the attention mechanism in computer vision~\cite{Dosovitskiy2020VIT}, attention has been utilized to process global context, which is fundamental to alleviate depth scale ambiguity~\cite{Bhat2020, Yang2021, Ranftl2021}.
State-of-the-art (SotA) methods typically involve convolutional networks~\cite{Eigen2014, Fu2018, Lee2019} or, since the advent of vision Transformer~\cite{Dosovitskiy2020VIT}, transformer architectures~\cite{Bhat2020, Yang2021, Ranftl2021, Yuan2022}. Most methods either impose geometric constraints on the image~\cite{Yin2019, Huynh2020, Long2021, Patil2022}, namely, planarity priors or explicitly discretize the continuous depth range~\cite{Fu2018, Bhat2020, Bhat2022}. The latter can be viewed as learning frontoparallel planes. These imposed priors inherently limit the expressiveness of the respective models, as they cannot model \emph{arbitrary} depth patterns, ubiquitous in real-world scenes.

We instead propose a more general depth estimation model, called \ourmodel, which does not explicitly impose any constraint on the final prediction. We design an \ourmodulename(\ourmodule) of the scene which is in principle depth-agnostic. Our assumption behind this \ourmodule is that each scene can be implicitly described by a set of concepts or patterns, such as objects, planes, edges, and perspectivity relationships. The specific training signal determines which patterns to learn (see \cref{fig:teaser}). 

We design a continuous-to-discrete bottleneck through which the information is passed in order to obtain such internal scene discretization, namely the underlying patterns. In the bottleneck, the scene feature space is partitioned via learnable and input-dependent quantizers, which in turn transfer the information onto the continuous output space. The \ourmodule bottleneck introduced in this work is a general concept and can be implemented in several ways. Our particular \ourmodule implementation employs attention-based operators, leading to an end-to-end trainable architecture and input-dependent framework. More specifically, we implement the continuous-to-discrete operation via ``transposed'' cross-attention, where transposed refers to applying $\mathrm{softmax}$ on the output dimension. This $\mathrm{softmax}$ formulation enforces the input features to be routed to the internal discrete representations (IDRs) in an exclusive fashion, thus defining an input-dependent soft clustering of the feature space. The discrete-to-continuous transformation is implemented via cross-attention. Supervision is only applied to the final output, without any assumptions or regularization on the IDRs. 

We test \ourmodel on multiple indoor and outdoor datasets and probe its robustness via zero-shot testing. As of today, there is too little variety in MDE benchmarks for the outdoor scenario, since the only established benchmark is KITTI~\cite{Geiger2012}. Moreover, we observe that all methods fail on outdoor zero-shot testing, suggesting that the KITTI dataset is not diverse enough and leads to overfitting, thus implying that it is not indicative of generalized performance. Hence, we find it compelling to establish a new benchmark setup for the MDE community by proposing two new train-test splits of more diverse and challenging high-quality outdoor datasets: Argoverse1.1~\cite{Chang2019} and DDAD~\cite{Guizilini2020}. 

Our main contributions are as follows: (i) we introduce the \ourmodulename module, a novel architectural component that adeptly represents a scene by combining underlying patterns; (ii) we show that it is a generalization of SotA methods involving depth ordinal regression~\cite{Fu2018, Bhat2020}; (iii) we propose splits of two raw outdoor datasets~\cite{Chang2019, Guizilini2020} with high-quality LiDAR measurements. We extensively test \ourmodel on six diverse datasets and, owing to the \ourmodule design, our model consistently outperforms SotA methods and presents better transferability. Moreover, we apply \ourmodel to surface normal estimation showing that the proposed module is general enough to tackle generic real-valued dense prediction tasks.

%-------------------------------------------------------------------------
%------------------------------------------------------------------------
\section{Related Work}
\label{sec:relwork}

\begin{figure*}[t]
    \centering
    \includegraphics[width=1\textwidth]{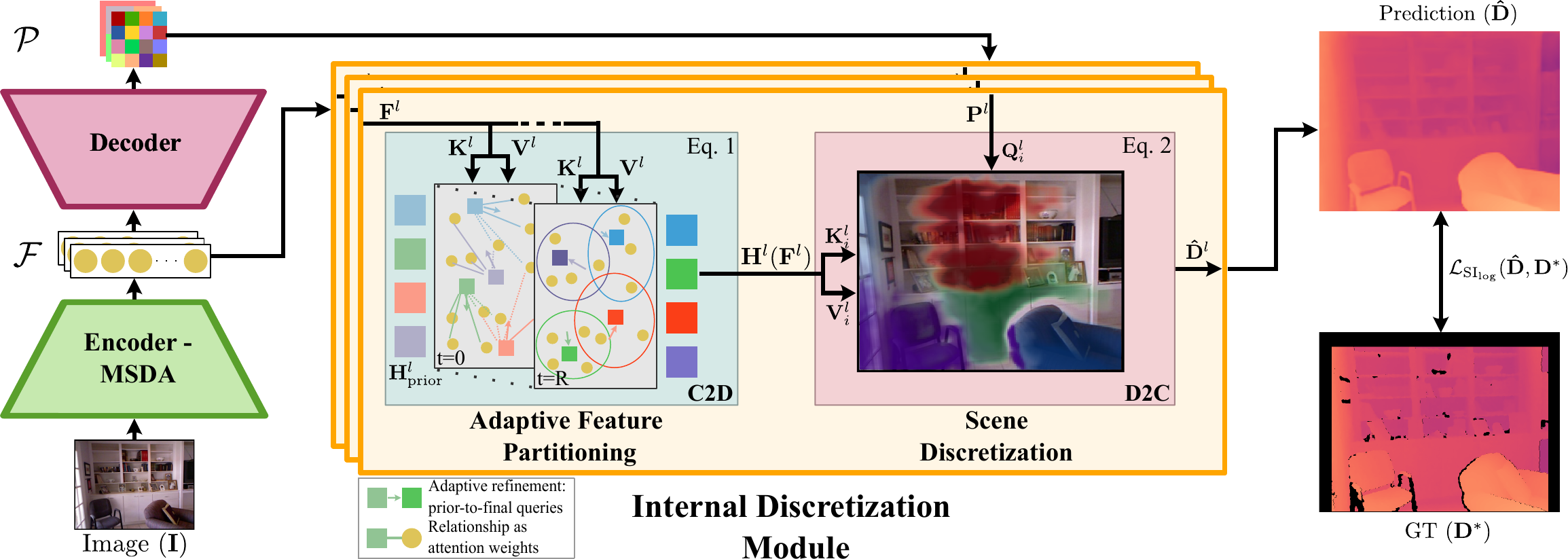}
    \vspace{-15pt}
    \caption{\textbf{Model Architecture.} The \ourmodulename Module imposes an information bottleneck via two consecutive stages: continuous-to-discrete (C2D) and discrete-to-continuous (D2C). The module processes multiple resolutions, \ie, $l \in \{1,2,3\}$, independently in parallel. The bottleneck embodies our assumption that a scene can be represented as a set of patterns. The C2D stage aggregates information, given a learnable prior ($\mathbf{H}^l_{\text{prior}}$), from the $l$-th resolution feature maps ($\mathbf{F}^l$) to a finite set of IDRs ($\mathbf{H}^l$). In particular, it learns how to define a partition function that is dependent on the input $\mathbf{F}^l$ via transposed cross-attention, as in \eqref{eqn:slotattn}. The second stage (D2C) transfers the IDRs on the original continuous space using layers of cross-attention as in \eqref{eqn:crossattn}, for sake of simplicity, we depict only a generic $i$-th layer. Cross-attention is guided by the similarity between decoded pixel embeddings ($\mathbf{P}^l$) and $\mathbf{H}^l$. The final prediction ($\mathbf{\hat{D}}$) is the fusion, \ie, mean, of the intermediate representations $\{\mathbf{\hat{D}}^l\}_{l=1}^3$.}
    \label{fig:model}
    \vspace{-15pt}
\end{figure*}

The supervised setting of MDE assumes that pixel-wise depth annotations are available at training time and depth inference is performed on single images. The coarse-to-fine network introduced in Eigen~\etal\cite{Eigen2014} is the cornerstone in MDE with end-to-end neural networks. The work established the optimization process via the Scale-Invariant log loss ($\mathrm{SI}_{\log}$). Since then, the three main directions evolve: new architectures, such as residual networks~\cite{Laina2016}, neural fields~\cite{Liu2015, Xu2018}, multi-scale fusion~\cite{Lee2018, Miangoleh2021}, transformers~\cite{Yang2021, Bhat2020, Yuan2022}; improved optimization schemes, such as reverse-Huber loss~\cite{Laina2016}, classification~\cite{Cao2016}, or ordinal regression~\cite{Fu2018, Bhat2020}; multi-task learning to leverage ancillary information from the related task, such as surface normals estimation or semantic segmentation~\cite{Eigen2014, Qi2018, Xu2018tasks}.

\noindent{}\textbf{Geometric priors} have been widely utilized in the literature, particularly the piecewise planarity prior~\cite{Gallup2010, Chauve2010, Szomoru2014}, serving as a proper real-world approximation. The geometric priors are usually incorporated by explicitly treating the image as a set of planes~\cite{Liu2018PlaneNet, Liu2018PlaneRCNN, Yu2019, Li2021}, using a plane-inducing loss~\cite{Yu2020p2net}, forcing pixels to attend to the planar representation of other pixels~\cite{Lee2019, Patil2022}, or imposing consistency with other tasks' output~\cite{Yin2019, Long2021, Bae2022}, like surface normals. Priors can focus on a more holistic scene representation by dividing the whole scene into 3D planes without dependence on intrinsic camera parameters~\cite{Yang2018, Zhang2020}, aiming at partitioning the scene into dominant depth planes. In contrast to geometric prior-based works, our method lifts any explicit geometric constraints on the scene. Instead, \ourmodel implicitly enforces the representation of scenes as a set of high-level patterns.

\noindent{}\textbf{Ordinal regression} methods~\cite{Fu2018, Bhat2020, Bhat2022} have proven to be a promising alternative to other geometry-driven approaches. The difference with classification models is that class ``values'' are learnable and are real numbers, thus the problem falls into the regression category. The typical SotA rationale is to explicitly discretize the continuous output depth range, rendering the approach similar to mask-based segmentation. Each of the scalar depth values is associated with a confidence mask which describes the probability of each pixel presenting such a depth value. Hence, SotA methods inherently assume that depth can be represented as a set of frontoparallel planes, that is, depth ``masks''. 

The main paradigm of ordinal regression methods is to first obtain hidden representations and scalar values of discrete depth values. The dot-product similarity between the feature maps and the depth representations is treated as logits and $\mathrm{softmax}$ is applied to extract confidence masks (in Fu~\etal~\cite{Fu2018} this degenerates to $\mathrm{argmax}$). Finally, the final  prediction is defined as the per-pixel weighted average of the discrete depth values, with the confidence values serving as the weights. \ourmodel draws connections with the idea of depth discretization. However, our \ourmodule module is designed to be depth-agnostic. The discretization occurs at the abstract level of \emph{internal} features from the \ourmodule bottleneck instead of the output depth level, unlike other methods.

\noindent{}\textbf{Iterative routing} is related to our ``transposed'' cross-attention. The first approach of this kind was Capsule Networks and their variants \cite{Sabour2017, Hinton2018capsule}. Some formulations~\cite{Tsai2020capsules, Locatello2020} employ different kinds of attention mechanisms. Our attention mechanism draws connections with~\cite{Locatello2020}. However, we do not allow permutation invariance, since our assumption is that each discrete representation internally describes a particular kind of pattern. In addition, we do not introduce any other architectural components such as gated recurrent units (GRU). In contrast to other methods, our attention is employed at a higher abstraction level, namely in the decoder.

%------------------------------------------------------------------------
%------------------------------------------------------------------------
\section{Method}
\label{sec:method}

We propose an \ourmodulename (\ourmodule) module, to discretize the internal feature representation of encoder-decoder network architectures. We hypothesize that the module can break down the scenes into coherent concepts without semantic supervision. This section will first describe the module design and then discuss the network architecture. \cref{sec:method_idd_afp} defines the formulation of ``transposed'' cross-attention outlined in \cref{sec:intro} and describes the main difference with previous formulations from \cref{sec:relwork}. Moreover, we derive in \cref{sec:method_idd_sd} how the \ourmodel formulation can be interpreted as a generalization of SotA ordinal regression methods by reframing their original formulation. Eventually, \cref{sec:method_arch} presents the optimization problem and the overall architecture.

%-------------------------------------------------------------------------
\subsection{\ourmodulename Module}
\label{sec:method_idd}
The \ourmodule module involves a continuous-discrete-continuous bottleneck composed of two main consecutive stages. The overall module is based on our hypothesis that scenes can be represented as a finite set of patterns. The first stage consists in a continuous-to-discrete component, namely soft-exclusive discretization of the feature space. More specifically, it enforces an input-dependent soft clustering on the feature maps in an image-to-set fashion. The second stage completes the internal scene discretization by mapping the learned IDRs onto the continuous output space. IDRs are not bounded to focus exclusively on depth planes but are allowed to represent any high-level pattern or concept, such as objects, relative locations, and planes in the 3D space. In contrast with SotA ordinal regression methods~\cite{Fu2018, Bhat2020, Bhat2022}, the IDRs are neither explicitly tied to depth values nor directly tied to the output. Moreover, our module operates at multiple intermediate resolutions and merges them only in the last layer. The overall architecture of \ourmodel, particularly our \ourmodule module, is shown in Fig.~\ref{fig:model}.

\subsubsection{Adaptive Feature Partitioning}
\label{sec:method_idd_afp}
% \noindent{}\textbf{Adaptive Feature Partitioning.}
The first stage of our \ourmodule module, \emph{Adaptive Feature Partitioning} (AFP), generates proper discrete representations ($\mathcal{H}\coloneqq\{\mathbf{H}^l\}^3_{l=1}$) that quantize the feature space ($\mathcal{F}\coloneqq\{\mathbf{F}^l\}^3_{l=1}$) at each resolution $l$. We drop the resolution superscript $l$ since resolutions are independently processed and only one generic resolution is treated here. \ourmodel does not simply learn fixed centroids, as in standard clustering, but rather learns how to define a partition function in an input-dependent fashion. More specifically, an iterative transposed cross-attention module is utilized. Given the specific input feature maps ($\mathbf{F}$), the iteration process refines (learnable) IDR priors ($\mathbf{H}_{\text{prior}}$) over $R$ iterations. 

More specifically, the term ``transposed'' refers to the different axis along which the $\mathrm{softmax}$ operation is applied, namely $\left[\mathrm{softmax}(\mathbf{K}\mathbf{Q}^{T})\right]^{T} \mathbf{V}$ instead of the canonical dot-product attention $ \mathrm{softmax}(\mathbf{Q} \mathbf{K}^{T}) \mathbf{V}$, with $\mathbf{Q}, \mathbf{K}, \mathbf{V}$ as query, key and value tensors, respectively. In particular, the tensors are obtained as projections of feature maps and IDR priors, $f_{\mathbf{Q}}(\mathbf{H}_{\text{prior}})$, $f_{\mathbf{K}}(\mathbf{F})$, $f_{\mathbf{V}}(\mathbf{F})$. The $t$-th iteration out of $R$ can be formulated as follows:
\begin{equation}
\label{eqn:slotattn}
    W_{ij}^{t} = \frac{\exp(\mathbf{k}_{i}^{T} \mathbf{q}_{j}^{t})}{\sum_{k=1}^{N} \exp(\mathbf{k}_{i}^{T} \mathbf{q}_{k}^{t})}, \mathbf{q}_{j}^{t+1} = \sum_{i=1}^{M} W_{ij}^{t}\mathbf{v}_i,
\end{equation}
where $\mathbf{q}_{j}, \mathbf{k}_{i}, \mathbf{v}_{i} \in \mathbb{R}^{C}$ are query, key and value respectively, $N$ is the number of IDRs, nameley, clusters, and $M$ is the number of pixels. The weights $W_{ij}$ may be normalized to 1 along the $i$ dimension to avoid vanishing or exploding quantities due to the summation of un-normalized distribution. 

The quantization stems from the inherent behavior of $\mathrm{softmax}$. In particular, $\mathrm{softmax}$ forces competition among outputs: one output can be large only to the detriment of others. Therefore, when fixing $i$, namely, given a feature, only a few attention weights ($W_{ij}$) may be significantly greater than zero. Hence, the content $\mathbf{v}_{i}$ is routed only to a few IDRs at the successive iteration. Feature maps are fixed during the process and weights are shared by design, thus $\{\mathbf{k}_{i}, \mathbf{v}_{i}\}_{i=1}^{M}$ are the same across iterations. The induced competition enforces a soft clustering of the input feature space, where the last-iteration IDR represents the actual partition function ($\mathbf{H}\coloneqq\mathbf{Q}^{R}$). The probabilities of belonging to one partition are the attention weights, namely $W^{R}_{ij}$ with $j$-th query fixed. Since attention weights are inherently dependent on the input, the specific partitioning also depends on the input and takes place at inference time. The entire process of AFP leads to (soft) mutually exclusive IDRs.

As far as the partitioning rationale is concerned, the proposed AFP draws connections with iterative routing methods described in \cref{sec:relwork}. However, important distinctions apply. First, IDRs are not randomly initialized as the ``slots'' in Locatello~\etal~\cite{Locatello2020} but present a learnable prior. Priors can be seen as learnable positional embeddings in the attention context, thus we do not allow a permutation-invariant set of representations. Moreover, non-adaptive partitioning can still take place via the learnable priors if the iterations are zero. Second, the overall architecture differs noticeably as described in \cref{sec:relwork}, and in addition, \ourmodel partitions feature space at the decoder level, corresponding to more abstract, high-level concepts, while the SotA formulations focus on clustering at an abstraction level close to the input image.
One possible alternative approach to obtaining the aforementioned IDRs is the well-known image-to-set proposed in DETR~\cite{Carion2020}, namely via classic cross-attention between representations and image feature maps. However, the corresponding representations might redundantly aggregate features, where the extreme corresponds to each output being the mean of the input. Studies~\cite{Gao2021, Sun2020} have shown that slow convergence in transformer-based architectures may be due to the non-localized context in cross-attention. The exclusiveness of the IDRs discourages the redundancy of information in different IDRs. We argue that exclusiveness allows the utilization of fewer representations (32 against the 256 utilized in \cite{Bhat2020} and \cite{Fu2018}), and can improve both the interpretability of what IDRs are responsible for and training convergence.

\subsubsection{Internal Scene Discretization}
\label{sec:method_idd_sd}
% \noindent{}\textbf{Scene Discretization.}
In the second stage of the \ourmodule module, \emph{Internal Scene Discretization} (ISD), the module ingests pixel embeddings ($\mathcal{P}\coloneqq\{\mathbf{P}^l\}^3_{l=1}$) from the decoder and IDRs $\mathcal{H}$ from the first stage, both at different resolutions $l$, as shown in \cref{fig:model}. Each discrete representation carries both the signature, as the \textit{key}, and the output-related content, as the \textit{value}, of the pattern it represents. The similarity between IDRs and pixel embeddings is computed in order to spatially localize in the continuous output space where to transfer the information of each IDR. We utilize the dot-product similarity function. 

Furthermore, the kind of information to transfer onto the final prediction is not constrained, as we never explicitly handle depth values, usually called bins, until the final output. Thus, the IDRs are completely free to carry generic high-level concepts (such as object-ness, relative positioning, and geometric structures). This approach is in stark contrast with SotA methods \cite{Fu2018, Bhat2020, Li2022DF, Bhat2022}, which explicitly constrain what the representations are about: scalar depth values. Instead, \ourmodel learns to generate unconstrained representations in an input-dependent fashion. The effective discretization of the scene occurs in the second stage thanks to the information transfer from the set of exclusive concepts ($\mathcal{H}$) from AFP to the continuous space defined by $\mathcal{P}$. We show that our method is not bounded to depth estimation, but can be applied to generic continuous dense tasks, for instance, surface normal estimation. Consequently, we argue that the training signal of the task at hand determines how to internally discretize the scene, rendering our \ourmodule module general and usable in settings other than depth estimation.

From a practical point of view, the whole second stage consists in cross-attention layers applied to IDRs and pixel embeddings. As described in \cref{sec:method_idd_afp}, we drop the resolution superscript $l$. After that, the final depth maps are projected onto the output space and the multi-resolution depth predictions are combined. The $i$-th layer is defined as:
\begin{equation}
\label{eqn:crossattn}
    \mathbf{D}_{i+1} = \mathrm{softmax}(\mathbf{Q}_{i} \mathbf{K}_{i}^{T}) \mathbf{V}_{i} + \mathbf{D}_{i},
\end{equation}
where $\mathbf{Q}_{i} = f_{Q_i}(\mathbf{P}) \in \mathbb{R}^{H\times{}W\times{}C}$, $\mathbf{P}$ are pixel embeddings with shape $(H,W)$, and $\mathbf{K}_{i}, \mathbf{V}_{i} \in \mathbb{R}^{N \times C}$ are the $N$ IDRs under linear transformations $f_{K_i}(\mathbf{H})$, $f_{V_i}(\mathbf{H})$. The term $\mathbf{Q}_{i} \mathbf{K}_{i}^{T}$ determines the spatial location for which each specific IDR is responsible, while $\mathbf{V}_{i}$ carries the semantic content to be transferred in the proper spatial locations.

Our approach constitutes a generalization of depth estimation methods that involve (hybrid) ordinal regression. As described in \cref{sec:relwork}, the common paradigm in ordinal regression methods is to explicitly discretize depth in a set of masks with a scalar depth value associated with it. Then, they predict the likelihood that each pixel belongs to such masks. Our change of paradigm stems from the reinterpretation of the mentioned ordinal regression pipeline which we translate into the following mathematical expression:
\begin{equation}
\label{eqn:explicit_disc}
    \mathbf{D} = \mathrm{softmax}(\mathbf{P} \mathbf{R}^{T} \mathbin{/} T) \mathbf{v},
\end{equation}
where $\mathbf{P}$ are the pixel embeddings at maximum resolution and $T$ is the $\mathrm{softmax}$ temperature. $\mathbf{v} \in \mathbb{R}^{N \times 1}$ are $N$ depth scalar values and $\mathbf{R} \in \mathbb{R}^{N \times (C-1)}$ are their hidden representations, both processed as a unique stacked tensor ($\mathbf{R} || \mathbf{v} \in \mathbb{R}^{N \times C}$). From the reformulation in \eqref{eqn:explicit_disc}, one can observe that \eqref{eqn:explicit_disc} is a degenerate case of \eqref{eqn:crossattn}. In particular, $f_{Q}$ degenerates to the identity function. $f_{K}$ and $f_{V}$ degenerate to selector functions: the former function selects up to the $C-1$ dimensions and the latter selects the last dimension only. Moreover, the hidden representations are refined pixel embeddings ($f(\mathbf{P}_i) = \mathbf{H}_i = \mathbf{R} || \mathbf{v}$), and $\mathbf{D}$ in \eqref{eqn:explicit_disc} is the final output, namely no multiple iterations are performed as in \eqref{eqn:crossattn}. The explicit entanglement between the semantic content of the hidden representations and the final output is due to hard-coding $\mathbf{v}$ as depth scalar values.

%-------------------------------------------------------------------------
\subsection{Network Architecture}
\label{sec:method_arch}
Our network described in \cref{fig:model} comprises first an encoder backbone, interchangeably convolutional or attention-based, producing features at different scales. The encoded features at different resolutions are refined, and information between resolutions is shared, both via four multi-scale deformable attention (MSDA) blocks~\cite{Zhu2021DefDETR}. The feature maps from MSDA at different scales are fed into the AFP module to extract IDRs ($\mathcal{H}$), and into the decoder to extract pixel embeddings in the continuous space ($\mathcal{P}$). Pixel embeddings at different resolutions are combined with the respective IDRs in the ISD stage of the \ourmodule module to extract the depth maps. The final depth prediction corresponds to the mean of the interpolated intermediate representations. 
% One may observe that the lowest resolution produces a depth structure of the overall scene, while higher resolution maps correct higher frequency structures. 
The optimization process is guided only by the established $\mathrm{SI_{log}}$ loss defined in~\cite{Eigen2014}, and no other regularization is exploited. $\mathrm{SI_{log}}$ is defined as:
\begin{equation}
    \begin{split}
    &\mathcal{L}_{\mathrm{SI_{log}}} (\epsilon) = \alpha \sqrt{\mathbb{V}[\epsilon] + \lambda \mathbb{E}^{2}[\epsilon]}\\
    &\text{with }\epsilon = \log(\hat{y}) - \log(y^*),
    \end{split}
    \label{eqn:eps_log}
\end{equation}
where $\hat{y}$ is the predicted depth and $y^*$ is the ground-truth (GT) value. $\mathbb{V}[\epsilon]$ and $\mathbb{E}[\epsilon]$ are computed as the empirical variance and expected value over all pixels, namely, $\{\epsilon_i\}_{i=1}^{N}$. $\mathbb{V}[\epsilon]$ is the purely scale-invariant loss, while $\mathbb{E}^{2}[\epsilon]$ fosters a proper scale. $\alpha$ and $\lambda$ are set to $10$ and $0.15$, as customary. 

%-------------------------------------------------------------------------
%-------------------------------------------------------------------------
\section{Experiments}
\label{sec:experiments}
\begin{figure}
    \centering
    \includegraphics[width=1\columnwidth]{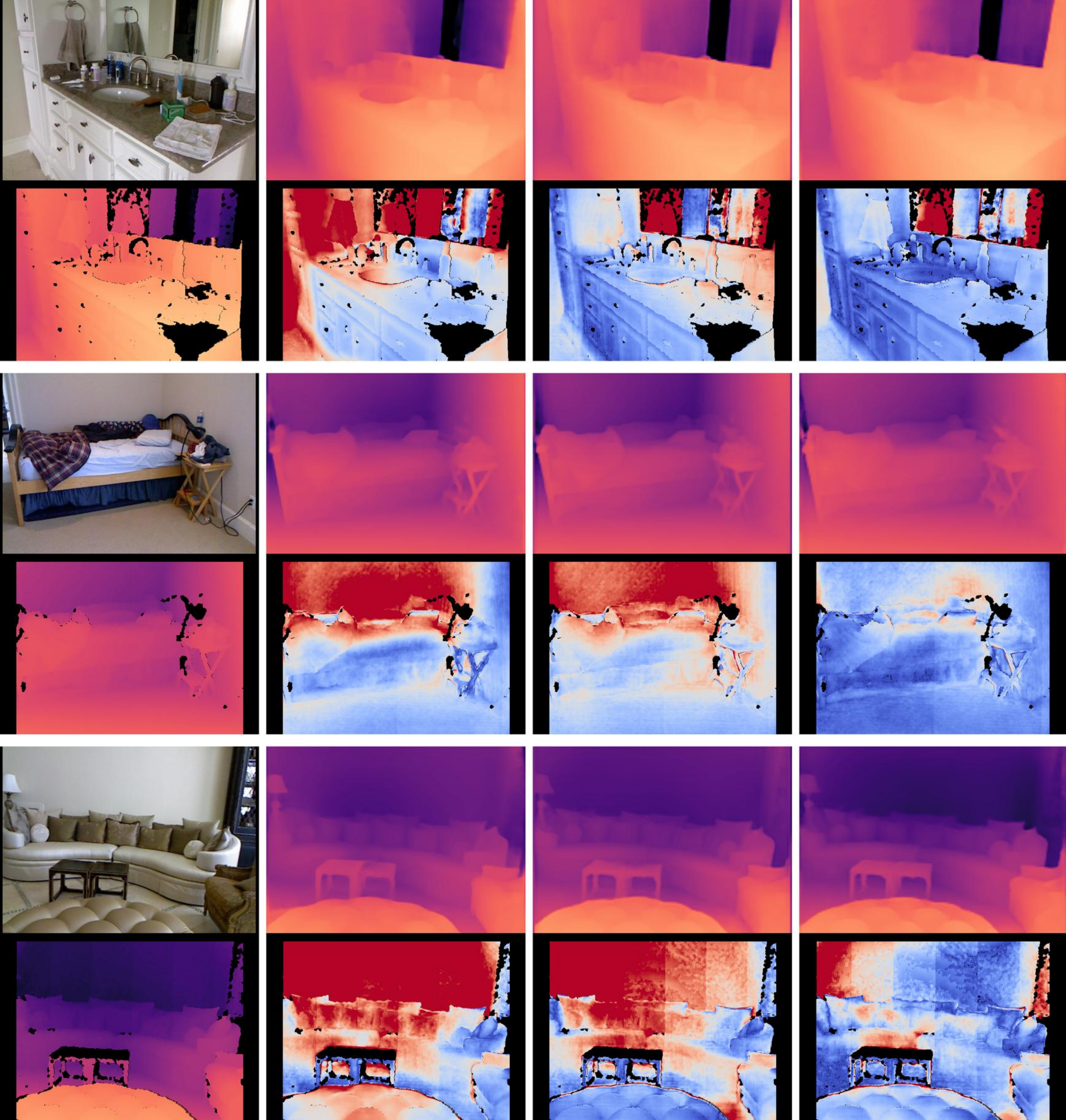}\\
    \footnotesize
    \begin{tabularx}{\columnwidth}{l}
        \hspace{0.02\columnwidth}\textbf{Image+GT}\hspace{0.09\columnwidth}\textbf{AdaBins}\cite{Bhat2020}\hspace{0.055\columnwidth}\textbf{NeWCRF}\cite{Yuan2022}\hspace{0.11\columnwidth}\textbf{Ours}
    \end{tabularx}\\
    \vspace{-5pt}
    \caption{\textbf{Qualitative results on NYU.} Each pair of consecutive rows corresponds to one test sample. Each odd row shows the input RGB image and depth predictions for the selected methods. Each even row shows GT depth and the prediction errors of the selected methods clipped at 0.5 meters. The error color map is \textit{coolwarm}: blue corresponds to lower error values and red to higher values.}
    \label{fig:nyu_results}
    \vspace{-10pt}
\end{figure}

%-------------------------------------------------------------------------
\subsection{Experimental Setup}
\label{sec:experiments_expsetup}
\subsubsection{Datasets}

\noindent{}\textbf{NYU-Depth V2.} NYU-Depth V2 (NYU)~\cite{Silberman:ECCV12} is a dataset consisting of 464 indoor scenes with RGB images and quasi-dense depth images with 640$\times{}$480 resolution. Our models are trained on the train-test split proposed by previous methods~\cite{Lee2019}, corresponding to 24,231 samples for training and 654 for testing. In addition to depth, the dataset provides surface normal data utilized for normal estimation. The train split used for normal estimation is the one proposed in~\cite{Yin2019}.

\noindent{}\textbf{Zero-shot testing datasets.} We evaluate the generalizability of indoor models on two indoor datasets which are not seen during training. The selected datasets are SUN-RGBD~\cite{Song2015} and DIODE-Indoor~\cite{Vasiljevic2019}. For both datasets, the resolution is reduced to match that of NYU, which is 640$\times{}$480.

\noindent{}\textbf{KITTI.} The KITTI dataset provides stereo images and corresponding Velodyne LiDAR scans of outdoor scenes captured from a moving vehicle~\cite{Geiger2012}. RGB and depth images have (mean) resolution of 1241$\times{}$376. The split proposed by~\cite{Eigen2014} (Eigen-split) with corrected depth is utilized as training and testing set, namely, 23,158 and 652 samples. The evaluation crop corresponds to the crop defined by~\cite{Garg2016}. All methods in \cref{sec:experiments_compsota} that have source code and pre-trained models available are re-evaluated on KITTI with the evaluation mask from~\cite{Garg2016} to have consistent results.

\noindent{}\textbf{Argoverse1.1 and DDAD.} We propose splits of two autonomous driving datasets, Argoverse1.1 (Argoverse)~\cite{Chang2019} and DDAD~\cite{Guizilini2020}, for depth estimation. Argoverse and DDAD are both outdoor datasets that provide 360$^{\circ}$ HD images and the corresponding LiDAR scans from moving vehicles. We pre-process the original datasets to extract depth maps and avoid redundancy. Training set scenes are sampled when the vehicle has been displaced by at least 2 meters from the previous sample. For the testing set scenes, we increase this threshold to 50 meters to further diminish redundancy. Our Argoverse split accounts for 21,672 training samples and 476 test samples, while DDAD for 18,380 training and 860 testing samples. Samples in Argoverse are taken from the 6 cameras covering the full 360$^{\circ}$ panorama. For DDAD, we exclude 2 out of the 6 cameras since they have more than 30\% pixels occluded by the camera capture system. 
We crop both RGB images and depth maps to have 1920$\times$870 resolution that is 180px and 210px cropped from the top for Argoverse and DDAD, respectively, to crop out a large portion of the sky and regions occluded by the ego-vehicle. For both datasets, we clip the maximum depth at 150m. 
% We will release code for the described preprocessing. These datasets are more varied than KITTI in terms of views, cities, and camera setups. 
% We hope that introducing in the common practice these two splits will allow the community to better validate models and to perform domain adaptation depth estimation in the real-to-real scenario.

\subsubsection{Implementation Details}
\begin{figure}[]
    \centering
    \includegraphics[width=\columnwidth]{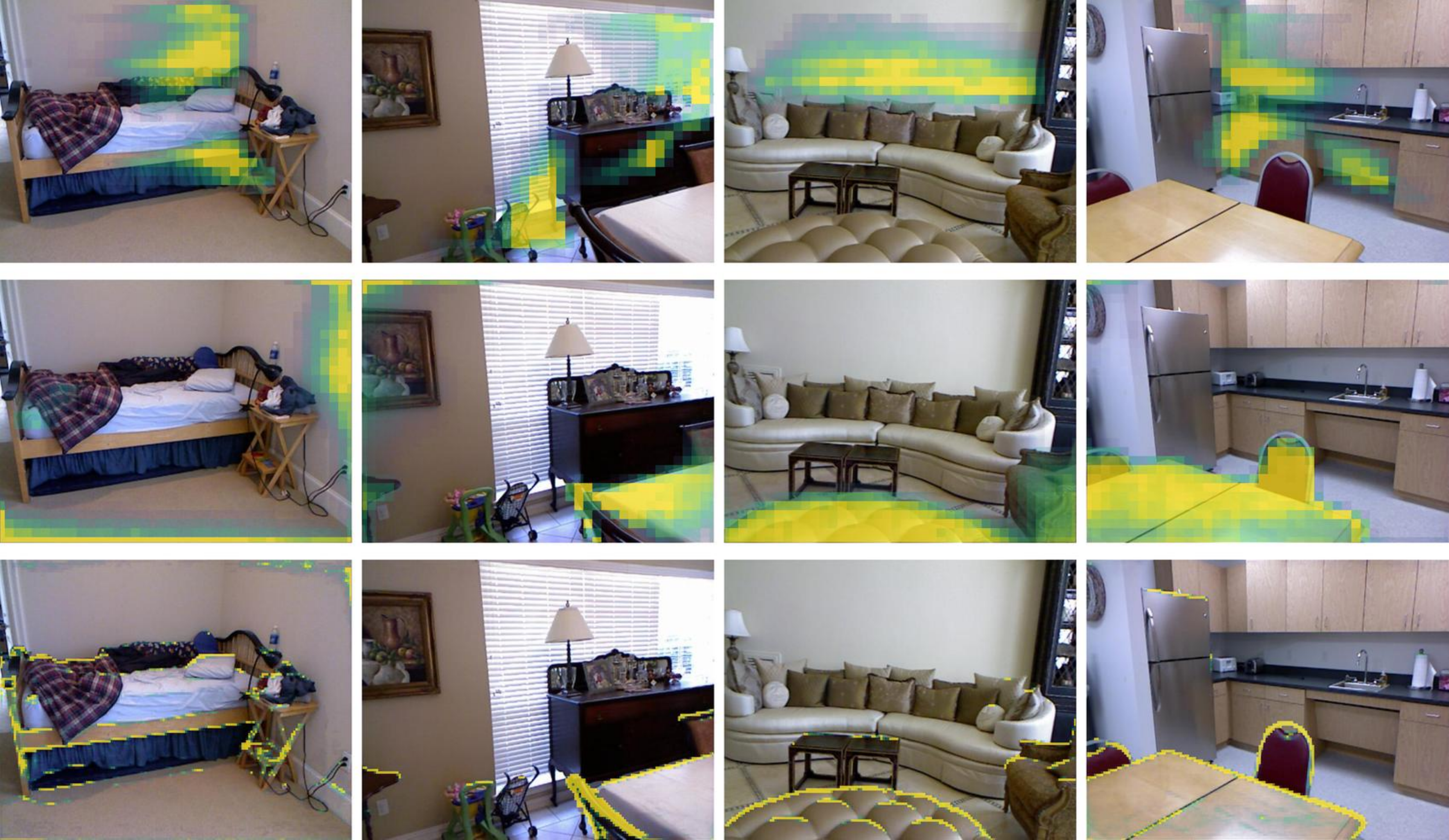}\\
    \vspace{-10pt}
    \caption{\textbf{Attention maps on NYU for three different IDRs.} Each row presents the attention map of a specific IDR for four test images. Each discrete representation focuses on a specific high-level concept. The first two rows pertain to IDRs at the lowest resolution while the last corresponds to the highest resolution. Best viewed on a screen and zoomed in.}
    \label{fig:nyu_attns}
    \vspace{-10pt}
\end{figure}
\noindent{}\textbf{Evaluation Details.} In all experiments, we do not exploit any test-time augmentations (TTA), camera parameters, or other tricks and regularizations, in contrast to many previous methods~\cite{Fu2018, Lee2019, Bhat2020, Patil2022, Yuan2022}. This provides a more challenging setup, which allows us to show the effectiveness of \ourmodel. 
As depth estimation metrics, we utilize root mean square error ($\mathrm{RMS}$) and its log variant ($\mathrm{RMS_{log}}$), absolute error in log-scale ($\mathrm{Log_{10}}$), absolute ($\mathrm{A.Rel}$) and squared ($\mathrm{S.rel}$) mean relative error, the percentage of inlier pixels ($\mathrm{\delta}_i$) with threshold $1.25^{i}$, and scale-invariant error in log-scale ($\mathrm{SI_{log}}$): $100 \sqrt{\mathrm{Var}(\epsilon_{\log})}$. The maximum depth for NYU and all zero-shot testing in indoor datasets, specifically SUN-RGBD and Diode Indoor, is set to 10m, while for KITTI it is set to 80m and for Argoverse and DDAD to 150m. Zero-shot testing is performed by evaluating a model trained on either KITTI or NYU and tested on either outdoor or indoor datasets, respectively, without additional fine-tuning. For surface normals estimation, the metrics are mean ($\mathrm{Mean}$) and median ($\mathrm{Med}$) absolute error, $\mathrm{RMS}$ angular error, and percentages of inlier pixels with thresholds at $11.5^{\circ}$, $22.5^{\circ}$, and $30^{\circ}$. GT-based mean depth rescaling is applied only on Diode Indoor for all methods since the dataset presents largely scale-equivariant scenes, such as plain walls with tiny details.

\noindent{}\textbf{Training Details.} We implement \ourmodel in PyTorch~\cite{pytorch}. For training, we use the AdamW~\cite{Loshchilov2017} optimizer ($\beta_1=0.9$, $\beta_2=0.999$) with an initial learning rate of $0.0002$ for every experiment, and weight decay set to $0.02$. As a scheduler, we exploit Cosine Annealing starting from 30\% of the training, with final learning rate of $0.00002$. We run 45k optimization iterations with a batch size of 16. 
% The number of locations of deformable attention is 4 for indoor datasets and 8 for outdoor ones, since outdoor images have twice as many pixels as indoor images. 
All backbones are initialized with weights from ImageNet-pretrained models. The augmentations include both geometric (random rotation and scale) and appearance (random brightness, gamma, saturation, hue shift) augmentations. The required training time amounts to 20 hours on 4 NVidia Titan RTX.

%-------------------------------------------------------------------------
\subsection{Comparison with the State of the Art}
\label{sec:experiments_compsota}

\begin{table}[]
    \centering
    \caption{\textbf{Comparison on NYU official test set.} R101: ResNet-101~\cite{He2015}, D161: DenseNet-161~\cite{Huang2016}, EB5: EfficientNet-B5~\cite{Tan2019}, HR48: HRNet-48~\cite{Wang2019}, DD22: DRN-D-22~\cite{Yu2017}, ViTB: ViT-B/16+Resnet-50~\cite{Dosovitskiy2020VIT}, MViT: EfficientNet-B5-AP~\cite{Xie2019}+MiniViT, Swin\{L, B, T\}: Swin-\{Large, Base, Tiny\}~\cite{Liu2021}. (\dag): ImageNet-22k~\cite{Deng2010} pretraining, (\ddag): non-standard training set, ($\ast$): in-house dataset pretraining, (\textsection): re-evaluated without GT-based rescaling.}%predictions rescaling via ground-truth means.}
    \vspace{-10pt}
    \resizebox{\columnwidth}{!}{
    \begin{tabular}{ll|ccc|ccc}
        \toprule
        \multirow{2}{*}{\textbf{Method}} & \multirow{2}{*}{\textbf{Encoder}} & $\boldsymbol{\delta_1}$ & $\boldsymbol{\delta_2}$ & $\boldsymbol{\delta_3}$ & $\mathbf{RMS}$ & $\mathbf{A.Rel}$ & $\mathbf{Log_{10}}$\\
        & & \multicolumn{3}{c|}{\textit{Higher is better}} & \multicolumn{3}{c}{\textit{Lower is better}}\\
        \toprule
        Eigen \etal~\cite{Eigen2014} & - & $0.769$ & $0.950$ & $0.988$ & $0.641$ & $0.158$  & $-$\\
        DORN~\cite{Fu2018} & R101 & $0.828$ & $0.965$ & $0.992$ & $0.509$ & $0.115$  & $0.051$\\
        VNL~\cite{Yin2019} & - & $0.875$ & $0.976$ & $0.994$ & $0.416$ & $0.108$ & $0.048$\\
        BTS~\cite{Lee2019} & D161 & $0.885$ & $0.978$ & $0.994$ & $0.392$ & $0.110$ & $0.047$\\
        AdaBins\textsuperscript{\ddag}~\cite{Bhat2020} & MViT & $0.903$ & $0.984$ & $0.997$ & $0.364$ & $0.103$ & $0.044$\\
        DAV~\cite{Huynh2020} & DD22 & $0.882$ & $0.980$ & $0.996$ & $0.412$ & $0.108$ & $-$ \\
        Long \etal~\cite{Long2021} & HR48 & $0.890$ & $0.982$ & $0.996$ & $0.377$ & $0.101$ & $0.044$ \\
        TransDepth~\cite{Yang2021} & ViTB & $0.900$ & $0.983$ & $0.996$ & $0.365$ & $0.106$ & $0.045$\\
        DPT*\cite{Ranftl2021} & ViTB  & $0.904$ & $0.988$ & $0.998$ & $0.357$ & $0.110$ & $0.045$\\
        % P3Depth\textsuperscript{\textsection}~\cite{Patil2022} & R101 & $0.898$ & $0.981$ & $0.996$ & $0.356$ & $0.104$ & $0.043$\\
        P3Depth\textsuperscript{\textsection}~\cite{Patil2022} & R101 & $0.830$ & $0.971$ & $0.995$ & $0.450$ & $0.130$ & $0.056$\\%no rescaling
        NeWCRF~\cite{Yuan2022} & SwinL\textsuperscript{\dag} & $0.922$ & $0.992$ & $0.998$ & $0.334$ & $0.095$ & $0.041$\\
        LocalBins\textsuperscript{\ddag}~\cite{Bhat2022} & MViT & $0.907$ & $0.987$ & $0.998$ & $0.357$ & $0.099$ & $0.042$\\
        \midrule
        \midrule
        Ours 
        & R101 & $0.892$ & $0.983$ & $0.995$ & $0.380$ & $0.109$ & $0.046$\\
        & EB5 & $0.903$ & $0.986$ & $0.997$ & $0.369$ & $0.104$ & $0.044$\\
        & SwinT & $0.894$ & $0.983$ & $0.996$ & $0.377$ & $0.109$ & $0.045$\\
        % & SwinS & $0.906$ & $0.986$ & $0.997$ & $0.354$ & $0.102$ & $0.043$\\
        & SwinB &  $0.926$ & $0.989$ & $ 0.997$ & $0.327$ & $0.091$ & $0.039$\\
        & SwinL\textsuperscript{\dag} & $\mathbf{0.940}$ & $\mathbf{0.993}$ & $\mathbf{0.999}$ & $\mathbf{0.313}$ & $\mathbf{0.086}$ & $\mathbf{0.037}$\\
        \bottomrule
    \end{tabular}}
    \label{tab:nyu_res}
    \vspace{-10pt}
\end{table}
\noindent{}\textbf{Indoor Datasets.} Results on NYU are presented in \Cref{tab:nyu_res}. The results show that we set the new state of the art on the benchmark, improving by more than 6\% on $\mathrm{RMS}$ and 9\% on $\mathrm{A.Rel}$ over the previous SotA. Moreover, results highlight how \ourmodel is more sample-efficient than other transformer-based architectures~\cite{Ranftl2021, Yang2021, Bhat2020, Yuan2022, Bhat2022} since we achieve better results even when employing smaller and less heavily pre-trained backbone architectures. In addition, results show a significant improvement in performance with our model instantiated with a full-convolutional backbone over other full-convolutional-based models~\cite{Eigen2014, Fu2018, Lee2019, Huynh2020, Patil2022}. \Cref{tab:nyu_zeroshot} presents zero-shot testing of NYU models on SUN-RGBD and Diode. In both cases, \ourmodel exhibits a compelling generalization performance, which we argue is due to implicitly learning the underlying patterns, namely, IDRs, of indoor scene structure via the \ourmodule module. 
\begin{table}[]
    \centering
    \caption{\textbf{Zero-shot testing of models trained on NYU.} All methods are trained on NYU and tested without further fine-tuning on the official validation set of SUN-RGBD and Diode Indoor.}
    \vspace{-10pt}
    \resizebox{\columnwidth}{!}{
    \begin{tabular}{llcccc}
        \toprule
        \textbf{Test set} & \textbf{Method} & $\boldsymbol{\delta_1} \uparrow$ & $\mathbf{RMS} \downarrow$ & $\mathbf{A.Rel} \downarrow$ & $\mathbf{SI_{log}} \downarrow$ \\
        \toprule
        SUN-RGBD
         & BTS~\cite{Lee2019} & $0.745$ & $0.502$ & $0.168$ & $14.25$\\
         & AdaBins~\cite{Bhat2020} & $0.768$ & $0.476$ & $0.155$ & $13.20$\\
         & P3Depth~\cite{Patil2022} & $0.698$ & $0.541$ & $0.178$ & $15.02$\\
         & NeWCRF~\cite{Yuan2022} & $0.799$ & $0.429$ & $0.150$ & $11.27$\\ 
         \cmidrule{2-6}
         & Ours & $\mathbf{0.838}$ & $\mathbf{0.387}$ & $\mathbf{0.128}$ & $\mathbf{10.91}$\\
        \midrule
        Diode  
         & BTS~\cite{Lee2019} & $0.705$ & $0.965$ & $0.211$ & $23.78$\\
         & AdaBins~\cite{Bhat2020} & $0.733$ & $0.872$ & $0.209$ & $22.54$\\
         & P3Depth~\cite{Patil2022} & $0.732$ & $0.877$ & $0.202$ & $22.16$\\
         & NeWCRF~\cite{Yuan2022} & $0.799$ & $0.769$ & $0.164$ & $18.69$\\
         \cmidrule{2-6}
         & Ours & $\mathbf{0.810}$ & $\mathbf{0.721}$ & $\mathbf{0.156}$ & $\mathbf{18.11}$\\
        \bottomrule
    \end{tabular}}
    \label{tab:nyu_zeroshot}
    \vspace{-10pt}
\end{table}

Qualitative results in \cref{fig:nyu_results} emphasize how the method excels in capturing the overall scene complexity. In particular, \ourmodel correctly captures discontinuities without depth over-excitation due to chromatic edges, such as the sink in row 1, and captures the right perspectivity between foreground and background depth planes such as between the bed (row 2) or sofa (row 3) and the walls behind. In addition, the model presents a reduced error around edges, even when compared to higher-resolution models such as~\cite{Bhat2020}. We argue that \ourmodel actually reasons at the pattern level, thus capturing better the structure of the scene. This is particularly appreciable in indoor scenes, since these are usually populated by a multitude of objects. This behavior is displayed in the attention maps of \cref{fig:nyu_attns}. \cref{fig:nyu_attns} shows how IDRs at lower resolution capture specific components, such as the relative position of the background (row 1) and foreground objects (row 2), while IDRs at higher resolution behave as depth refiners, attending typically to high-frequency features, such as upper (row 3) or lower borders of objects. It is worth noting that an IDR attends to the image borders when the particular concept it looks for is not present in the image. That is, the borders are the last resort in which the IDR tries to find its corresponding pattern (\eg, row 2, col.~1).

\begin{figure*}[]
    \centering
    \includegraphics[width=1\textwidth]{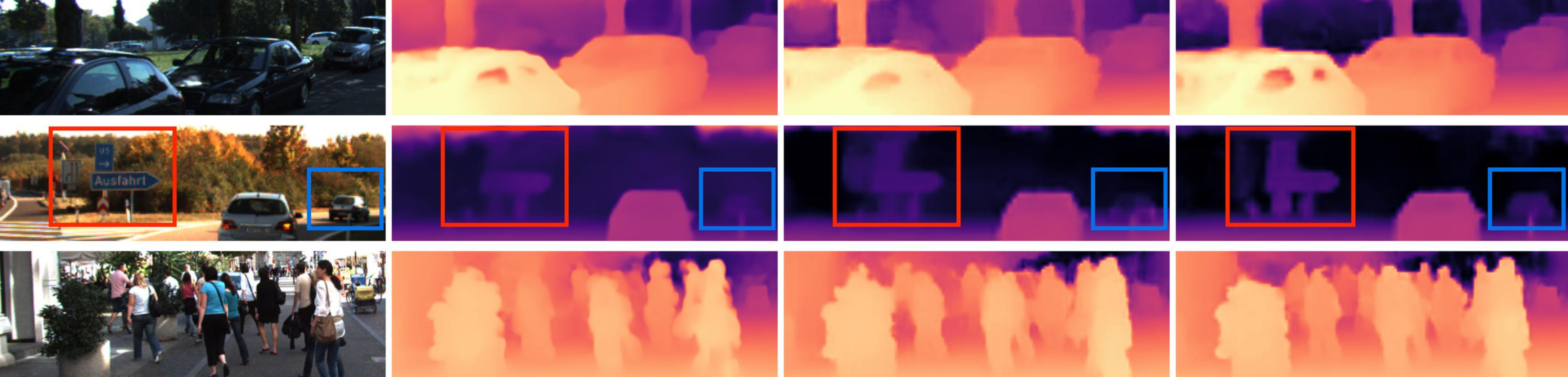}\\
    \footnotesize
    \begin{tabularx}{\textwidth}{l}
        \hspace{0.085\textwidth}\textbf{Image}\hspace{0.2\textwidth}\textbf{AdaBins}\cite{Bhat2020}\hspace{0.16\textwidth}\textbf{NeWCRF}\cite{Yuan2022}\hspace{0.18\textwidth}\textbf{Ours}
    \end{tabularx}\\
    \vspace{-5pt}
    \caption{\textbf{Qualitative results on KITTI.} Three zoomed-in crops of different test images are shown.
    % The colormap is \textit{magma reverse}, where black corresponds to 80 meters. 
    The comparisons show the ability of \ourmodel to capture small details, proper background transition, and fine-grained variations in, \eg, crowded scenes. Best viewed on a screen.
    % AdaBins works at double the resolution of other methods, this leads to a minor pixelation effect and, only visually, more appealing depth maps from AdaBins.
    }
    \label{fig:kitti_results}
    \vspace{-10pt}
\end{figure*}

\begin{table}[]
    \caption{\textbf{Comparison on KITTI Eigen-split test set.} Models without $\delta_{0.5}$ have implementation (partially) unavailable. R101: ResNet-101~\cite{He2015}, D161: DenseNet-161~\cite{Huang2016}, EB5: EfficientNet-B5\cite{Tan2019}, ViTB: ViT-B/16+Resnet-50~\cite{Dosovitskiy2020VIT}, MViT: EfficientNet-B5-AP~\cite{Xie2019}+MiniViT, Swin\{L, B, T\}: Swin-\{Large, Base, Tiny\}~\cite{Liu2021}. (\dag): ImageNet-22k~\cite{Deng2010} pretraining, (\ddag): non-standard training set, ($\ast$): in-house dataset pretraining, (\textsection): re-evaluated without GT-based rescaling.}%predictions rescaling via ground-truth means.}
    \vspace{-10pt}
    \resizebox{\columnwidth}{!}{
    \begin{tabular}{ll|ccc|cccc}
        \toprule
        \multirow{2}{*}{\textbf{Method}} & \multirow{2}{*}{\textbf{Encoder}} & $\boldsymbol{\delta_{0.5}}$ & $\boldsymbol{\delta_1}$ & $\boldsymbol{\delta_2}$ & $\mathbf{RMS}$ & $\mathbf{RMS_{log}}$ & $\mathbf{A.Rel}$ & $\mathbf{S.Rel}$\\
        & & \multicolumn{3}{c|}{\textit{Higher is better}} & \multicolumn{4}{c}{\textit{Lower is better}}\\
        \toprule
        Eigen\etal~\cite{Eigen2014} & $-$ & $-$ & $0.692$ & $0.899$ & $7.156$ & $0.270$ & $0.190$ & $1.515$\\
        DORN~\cite{Fu2018} & R101 & $-$ & $0.932$ & $0.984$ & $2.727$ & $0.120$ & $0.072$ & $0.307$\\
        BTS~\cite{Lee2019} & D161 & $0.870$ & $0.964$ & $0.995$ & $2.459$ & $0.090$ & $0.057$ & $0.199$\\
        AdaBins\textsuperscript{\ddag}~\cite{Bhat2020} & MViT & $0.868$ & $0.964$ & $0.995$ & $2.360$ &  $0.088$ & $0.058$ & $0.198$\\
        TransDepth~\cite{Yang2021} & ViTB & $-$ & $0.956$ & $0.994$ & $2.755$ & $0.098$ & $0.064$ & $0.252$\\
        DPT*~\cite{Ranftl2021} & ViTB & $0.865$ & $0.965$ & $0.996$ & $2.315$ & $0.088$ & $0.059$ & $0.190$\\
        % P3Depth\textsuperscript{\textsection}~\cite{Patil2022} & R101 & $0.873$ & $0.967$ & $0.995$ & $2.431$ & $0.089$ & $0.056$ & $0.189$\\
        P3Depth\textsuperscript{\textsection}~\cite{Patil2022} & R101 & $0.852$ & $0.959$ & $0.994$ & $2.519$ & $0.095$ & $0.060$ & $0.206$\\%no rescaling
        NeWCRF~\cite{Yuan2022} & SwinL\textsuperscript{\dag} & $0.887$ & $0.974$ & $\mathbf{0.997}$ & $2.129$ & $0.079$ & $0.052$ & $0.155$\\
        \midrule
        \midrule
        Ours
        & R101 & $0.860$ & $0.965$ & $0.996$ & $2.362$ & $0.090$ & $0.059$ & $0.197$\\
        & EB5 & $0.852$ & $0.963$ & $0.994$ & $2.510$ & $0.094$ & $0.063$ & $0.223$\\
        & SwinT & $0.870$ & $0.968$ & $0.996$ & $2.291$ & $0.087$ & $0.058$ & $0.184$\\
        % & SwinS & $0.880$ & $0.972$ & $\mathbf{0.997}$ & $2.219$ & $0.083$ & $0.054$ & $0.166$\\
        & SwinB & $0.885$ & $0.974$ & $\mathbf{0.997}$ & $2.149$ & $0.081$ & $0.054$ & $ 0.159$\\
        & SwinL\textsuperscript{\dag} & $\mathbf{0.896}$ & $\mathbf{0.977}$ & $\mathbf{0.997}$ & $\mathbf{2.067}$ & $\mathbf{0.077}$ & $\mathbf{0.050}$ & $\mathbf{0.145}$\\
        \bottomrule
    \end{tabular}}
    \label{tab:kitti_res}
    \vspace{-0pt}
\end{table}
\noindent{}\textbf{Outdoor Datasets.} Results on KITTI in \Cref{tab:kitti_res} demonstrate that \ourmodel sets the new SotA for this primary outdoor dataset, improving by more than 3\% in $\mathrm{RMS}$ and by 0.9\% in $\mathrm{\delta_{0.5}}$ over the previous SotA. However, KITTI results present saturated metrics. For instance, $\mathrm{\delta_3}$ is not reported since every method scores $>0.99$, with recent ones scoring $0.999$. Therefore, we propose to utilize the metric $\mathrm{\delta_{0.5}}$, to better convey meaningful evaluation information. In addition, \ourmodel performs remarkably well on the highly competitive official KITTI benchmark, ranking 3\textsuperscript{rd} among all methods and 1\textsuperscript{st} among all published MDE methods.

Moreover, \Cref{tab:ddad_argo_results} shows the results of methods trained and evaluated on the splits from Argoverse and DDAD proposed in this work. All methods have been trained with the same architecture and pipeline utilized for training on KITTI. We argue that the high degree of sparseness in GT of the two proposed datasets, in contrast to KITTI, deeply affects windowed methods such as~\cite{Bhat2020, Yuan2022}. Qualitative results in \cref{fig:kitti_results} suggest that the scene level discretization leads to retaining small objects and sharp transitions between foreground objects and background: background in row 1, and boxes in row 2. These results show the better ability of \ourmodel to capture fine-grained depth variations on close-by and similar objects, including crowd in row 3. Zero-shot testing from KITTI to DDAD and Argoverse are presented in Supplement.
% \Cref{tab:kitti_zeroshot}.

\begin{table}[]
    \centering
    \caption{\textbf{Comparison on Argoverse and DDAD proposed splits.} Comparison of performance of methods trained on either Argoverse or DDAD and tested on the same dataset.}
    \vspace{-10pt}
    \resizebox{\columnwidth}{!}{
    \begin{tabular}{ll|ccc|cccc}
        \toprule
        \multirow{2}{*}{\textbf{Dataset}} & \multirow{2}{*}{\textbf{Method}} & $\boldsymbol{\delta_{1}}$ & $\boldsymbol{\delta_2}$ & $\boldsymbol{\delta_3}$ & $\mathbf{RMS}$ & $\mathbf{RMS_{log}}$ & $\mathbf{A.Rel}$ & $\mathbf{S.Rel}$\\
        & & \multicolumn{3}{c|}{\textit{Higher is better}} & \multicolumn{4}{c}{\textit{Lower is better}}\\
        \toprule
        Argoverse
         & BTS~\cite{Lee2019} & $0.780$ & $0.908$ & $0.954$ & $8.319$ & $0.267$ & $0.186$ & $2.56$\\
         & AdaBins~\cite{Bhat2020} & $0.750$ & $0.901$ & $0.952$ & $8.686$ & $0.278$ & $0.195$ & $2.36$\\
         & NeWCRF~\cite{Yuan2022} & $0.707$ & $0.871$ & $0.939$ & $9.437$ & $0.321$ & $0.232$ & $3.23$\\
         \cmidrule{2-9}
         & Ours & $\mathbf{0.821}$ & $\mathbf{0.923}$ & $\mathbf{0.960}$ & $\mathbf{7.567}$ & $\mathbf{0.243}$ & $\mathbf{0.163}$ & $\mathbf{2.22}$\\
        \midrule
        DDAD
         & BTS~\cite{Lee2019} & $0.757$ & $0.913$ & $0.962$ & $10.11$ & $0.251$ & $0.186$ & $2.27$\\
         & AdaBins~\cite{Bhat2020} & $0.748$ & $0.912$ & $0.962$ & $10.24$ & $0.255$ & $0.201$ & $2.30$\\
         & NeWCRF~\cite{Yuan2022} & $0.702$ & $0.881$ & $0.951$ & $10.98$ & $0.271$ & $0.219$ & $2.83$\\
         \cmidrule{2-9}
         & Ours & $\mathbf{0.809}$ & $\mathbf{0.934}$ & $\mathbf{0.971}$ & $\mathbf{8.989}$ & $\mathbf{0.221}$ & $\mathbf{0.163}$ & $\mathbf{1.85}$\\
        \bottomrule
    \end{tabular}}
    \label{tab:ddad_argo_results}
    \vspace{-10pt}
\end{table}

\noindent{}\textbf{Surface Normals Estimation.} We emphasize that the proposed method has more general applications by testing \ourmodel on a different continuous dense prediction task such as surface normals estimation. Results in \Cref{tab:nyu_normals_res} evidence that we set the new state of the art on surface normals estimation. It is worth mentioning that all other methods are specifically designed for normals estimation, while we keep the same architecture and framework from indoor depth estimation.
\begin{table}[]
    \centering
    \caption{\textbf{Comparison of surface normals estimation methods on NYU official test set.} \ourmodel architecture and training pipeline is the same as the one utilized for indoor depth estimation.}
    \vspace{-10pt}
    \resizebox{\columnwidth}{!}{
    \begin{tabular}{l|ccc|ccc}
        \toprule
        \multirow{2}{*}{\textbf{Method}} & $\boldsymbol{11.5^{\circ}}$ & $\boldsymbol{22.5^{\circ}}$ & $\boldsymbol{30^{\circ}}$ & $\mathbf{RMS}$ & $\mathbf{Mean}$ & $\mathbf{Med}$\\
        & \multicolumn{3}{c|}{\textit{Higher is better}} & \multicolumn{3}{c}{\textit{Lower is better}}\\
        \toprule
        % Eigen \etal\cite{Eigen2014} & $0.444$ & $0.672$ & $0.759$ & $-$ & $20.9$ & $13.2$\\
        % SkipNet & $0.479$ & $0.700$ & $0.778$ & $28.2$ & $19.8$ & $12.0$\\
        SURGE~\cite{Wang2016} & $0.473$ & $0.689$ & $0.766$ & $-$ & $20.6$ &  $12.2$\\
        GeoNet~\cite{Qi2018} & $0.484$ & $0.484$ & $0.795$ & $26.9$ & $19.0$ & $11.8$\\
        PAP~\cite{Zhang2019} & $0.488$ & $0.722$ & $0.798$ & $25.5$ & $18.6$ & $11.7$\\
        GeoNet++~\cite{Qi2022} & $0.502$ & $0.732$ & $0.807$ & $26.7$ & $18.5$ & $11.2$\\
        Bae \etal~\cite{Bae2021} & $0.622$ & $0.793$ & $0.852$ & $23.5$ & $14.9$ & $7.5$\\
        \midrule
        \midrule
        Ours & $\mathbf{0.638}$ & $\mathbf{0.798}$ & $\mathbf{0.856}$ & $\mathbf{22.8}$ & $\mathbf{14.6}$ & $\mathbf{7.3}$\\
        \bottomrule
    \end{tabular}}
    \label{tab:nyu_normals_res}
    \vspace{-10pt}
\end{table}

\subsection{Ablation study}
\label{sec:experiments_ablations}
The importance of each component introduced in \ourmodel is evaluated by ablating the method in \Cref{tab:ablations}. 

\noindent{}\textbf{Depth Discretization.} Internal scene discretization provides a clear improvement over its explicit counterpart (row 3 vs.\ 2), which is already beneficial in terms of robustness. Adding the MSDA module on top of explicit discretization (row 5) recovers part of the performance gap between the latter and our full method (row 8). We argue that MSDA recovers a better scene scale by refining feature maps at different scales at once, which is helpful for higher-resolution feature maps.

\noindent{}\textbf{Component Interactions.} Using either the MSDA module or the AFP module together with internal scene discretization results in similar performance (rows 4 and 6). We argue that the two modules are complementary, and they synergize when combined (row 8). The complementarity can be explained as follows: in the former scenario (row 4), MSDA preemptively refines feature maps to be partitioned by the non-adaptive clustering, that is, by the IDR priors described in \cref{sec:method}, while on latter one (row 6), AFP allows the IDRs to adapt themselves to partition the unrefined feature space properly. Row 7 shows that the architecture closer to the one in~\cite{Locatello2020}, particularly random initialization, hurts performance since the internal representations do not embody any domain-specific prior information.

\begin{table}[]
    \centering
    \caption{\textbf{Ablation of \ourmodel.} EDD: Explicit Depth Discretization \cite{Fu2018, Bhat2020}, ISD: Internal Scene discretization, AFP: Adaptive Feature Partitioning, MSDA: MultiScale Deformable Attention. The EDD module, used in SotA methods, and our ISD module are mutually exclusive. AFP with (\cmark$_{\textbf{R}}$) refers to random initialization of IDRs and architecture similar to~\cite{Locatello2020}. The last row corresponds to our complete \ourmodel model.}
    \vspace{-10pt}
    \resizebox{\columnwidth}{!}{
    \begin{tabular}{ccccc|ccc}
        \toprule
        & \textbf{EDD} & \textbf{ISD} & \textbf{AFP} & \textbf{MSDA} & $\boldsymbol{\delta_1} \uparrow$ & $\mathbf{RMS} \downarrow$ & $\mathbf{A.Rel} \downarrow$\\
        \toprule    
        1 & \xmark & \xmark & \xmark & \xmark & $0.890$ & $0.370$ & $0.104$\\
        2 & \cmark & \xmark & \xmark & \xmark & $0.905$ & $0.367$ & $0.102$\\
        3 & \xmark & \cmark & \xmark & \xmark & $0.919$ & $0.340$ & $0.096$\\
        4 & \xmark & \cmark & \cmark & \xmark & $0.931$ & $0.319$ & $0.091$\\
        \midrule
        5 & \cmark & \xmark & \xmark & \cmark & $0.931$ & $0.326$ & $0.091$\\
        6 & \xmark & \cmark & \xmark & \cmark & $0.934$ & $0.319$ & $0.088$\\
        \midrule
        7 & \xmark & \cmark & \cmark$_\textbf{R}$ & \cmark & $0.930$ & $0.319$ & $0.089$\\
        8 & \xmark & \cmark & \cmark & \cmark & $0.940$ & $0.313$ & $0.086$\\
        \bottomrule
    \end{tabular}
    }
    \label{tab:ablations}
    \vspace{-10pt}
\end{table}
%-------------------------------------------------------------------------
%------------------------------------------------------------------------
\section{Conclusion}
We have introduced a new module, called \ourmodulename, for MDE. The module represents the assumption that scenes can be represented as a finite set of patterns. Hence, \ourmodel leverages an internally discretized representation of the scene that is enforced via a continuous-discrete-continuous bottleneck, namely \ourmodule module. We have validated the proposed method, without any TTA or tricks, on the primary indoor and outdoor benchmarks for MDE, and have set the new state of the art among supervised approaches. Results showed that learning the underlying patterns, while not imposing any explicit constraints or regularization on the output, is beneficial for performance and generalization. \ourmodel also works out-of-the-box for normal estimation, beating all specialized SotA methods. In addition, we propose two new challenging outdoor dataset splits, aiming to benefit the community with more general and diverse benchmarks.

\vfill
\noindent{}\textbf{Acknowledgment.} This work is funded by Toyota Motor Europe via the research project TRACE-Z\"urich.

%% file: IDiscAppendix.tex
\def\thesection{\Alph{section}}
\setcounter{section}{0}
\setcounter{figure}{5}
\setcounter{table}{6}

\section{Results}
\begin{figure*}[h]
    \centering
    \includegraphics[width=\textwidth]{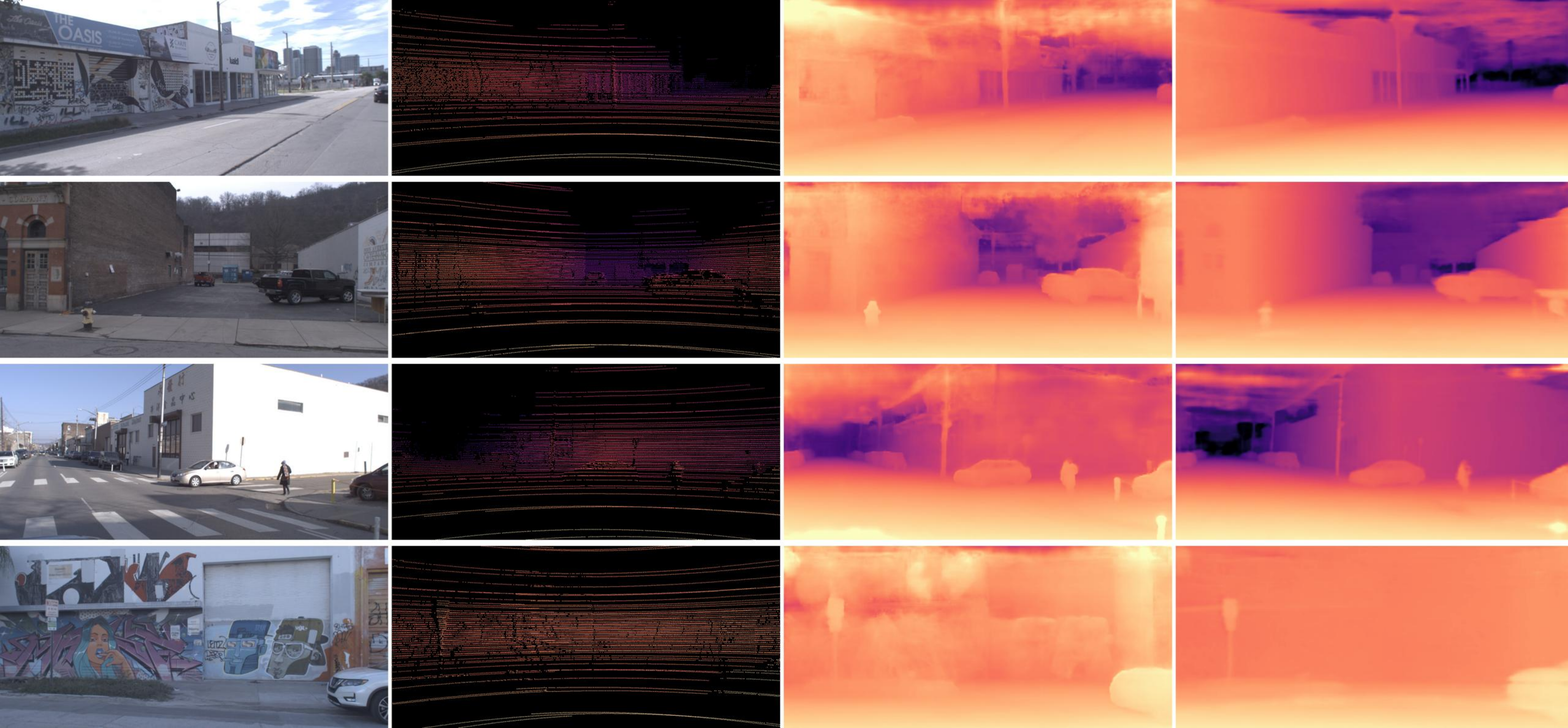}\\
    \footnotesize
    \begin{tabularx}{\textwidth}{l}
        \hspace{0.09\textwidth}\textbf{Image}\hspace{0.22\textwidth}\textbf{GT}\hspace{0.17\textwidth}\textbf{Ours (w/ zero-shot)}\hspace{0.17\textwidth}\textbf{Ours}
    \end{tabularx}\\
    \vspace{-5pt}
    \caption{\textbf{Qualitative results on Argoverse.} Each row corresponds to one zero-shot test sample from Argoverse. The third column displays the prediction of \ourmodel trained on KITTI and tested on Argoverse, while the fourth column corresponds to a model trained and tested on Argoverse.}
    \label{fig:argo}
\end{figure*}

\begin{figure*}[h]
    \centering
    \includegraphics[width=\textwidth]{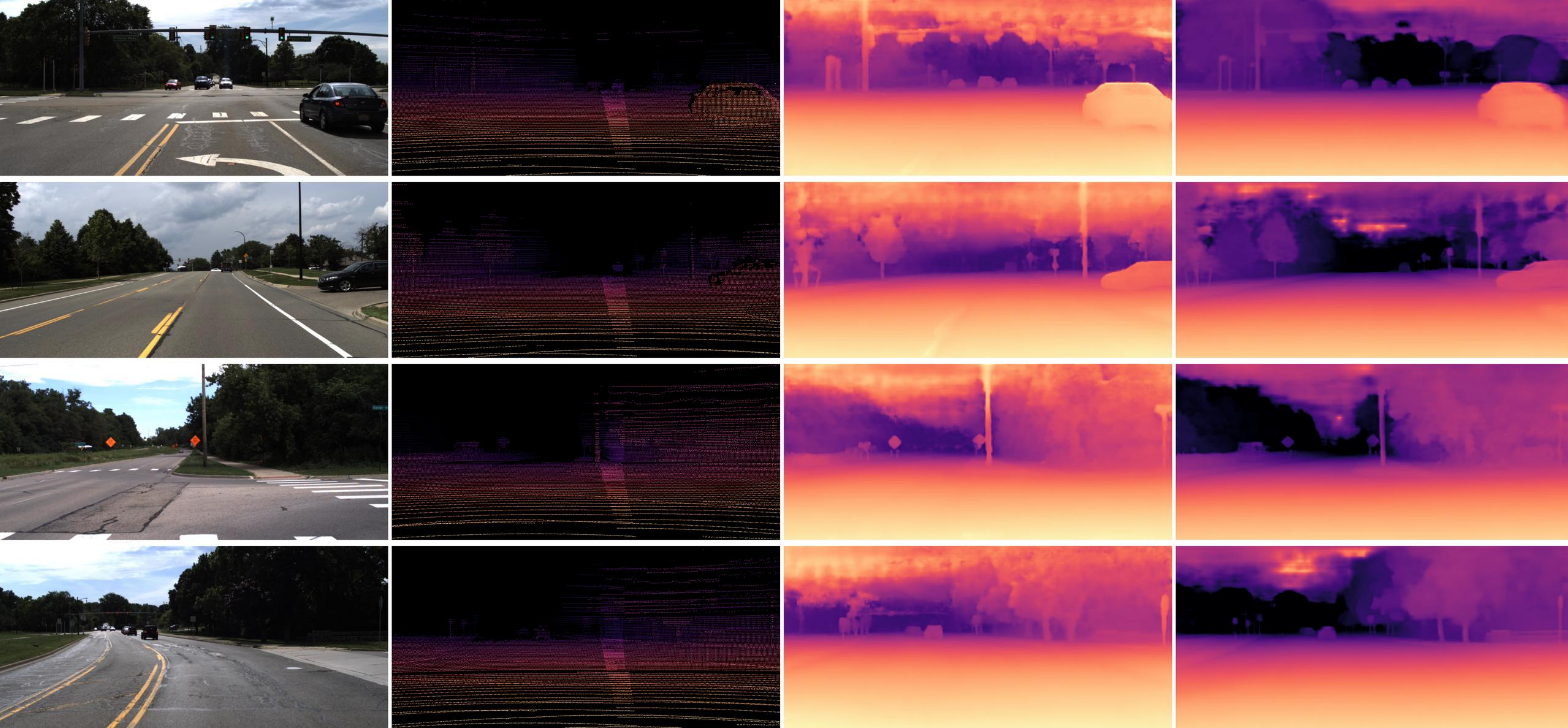}\\
    \footnotesize
    \begin{tabularx}{\textwidth}{l}
        \hspace{0.09\textwidth}\textbf{Image}\hspace{0.22\textwidth}\textbf{GT}\hspace{0.18\textwidth}\textbf{Ours (zero-shot)}\hspace{0.16\textwidth}\textbf{Ours (sup.)}
    \end{tabularx}\\
    \vspace{-5pt}
    \caption{\textbf{Qualitative results on DDAD.} Each row corresponds to one zero-shot test sample from DDAD. The third column displays the prediction of \ourmodel trained on KITTI and tested on DDAD, while the fourth corresponds column to a model trained and tested on DDAD.}
    \label{fig:ddad}
\end{figure*}

\noindent{}\textbf{Outdoor zero-shot.} We present in \Cref{tab:kitti_zeroshot} the results of models pre-trained on KITTI Eigen-split~\cite{Eigen2014} and tested on Argoverse~\cite{Chang2019}  and DDAD~\cite{Guizilini2020} test split we proposed in this work. The zero-shot results clearly demonstrate how every model tends to perform poorly when trained on KITTI and tested on a different domain. However, \ourmodel is able to almost double the performance when directly trained on either Argoverse or DDAD. This suggests that KITTI is not indicative of generalization performance. This investigation leads us to realize the need for more diversity in the outdoor scenario. We address the problem by proposing new dataset splits to train and validate models on. \cref{fig:argo} shows how models fail completely when predicting unseen scenario, \eg, graffiti on a flat wall. In addition, \cref{fig:ddad} displays how models under-scale depth when testing on domains with a typical object size, \ie, DDAD in the United States, larger than that of the training set, \ie, KITTI in Germany.

\begin{table}[b]
    \vspace{-10pt}
    \centering
    \caption{\textbf{Zero-shot testing of models trained on KITTI Eigen-split.} Comparison of performance when methods are trained on KITTI Eigen-split and tested, without further fine-tuning, on the splits of Argoverse and DDAD introduced in this work.}
    \vspace{-10pt}
    \resizebox{\columnwidth}{!}{
    \begin{tabular}{llcccc}
        \toprule
        \textbf{Test set} & \textbf{Method} & $\boldsymbol{\delta_1} \uparrow$ & $\mathbf{RMS} \downarrow$ & $\mathbf{A.Rel} \downarrow$ & $\mathbf{SI_{log}} \downarrow$ \\
        \toprule
        Argoverse
         & BTS~\cite{Lee2019} & $0.307$ & $15.98$ & $0.383$ & $51.80$\\
         & AdaBins~\cite{Bhat2020} & $0.383$ & $17.07$ & $0.350$ & $52.33$\\
         & P3Depth~\cite{Patil2022} & $0.277$ & $17.97$ & $0.376$ & $44.09$\\
         & NeWCRF~\cite{Yuan2022} & $0.311$ & $15.75$ & $0.370$ & $46.77$\\
         \cmidrule{2-6}
         & Ours & $\mathbf{0.560}$ & $\mathbf{12.18}$ & $\mathbf{0.269}$ & $\mathbf{33.35}$\\
        \midrule
        DDAD
         & BTS~\cite{Lee2019} & $\mathbf{0.399}$ & $16.19$ & $0.350$ & $40.51$\\
         & AdaBins~\cite{Bhat2020} & $0.282$ & $18.36$ & $0.433$ & $50.71$\\
         & P3Depth~\cite{Patil2022} & $0.397$ & $17.83$ & $\mathbf{0.330}$ & $39.00$\\
         & NeWCRF~\cite{Yuan2022} & $0.343$ & $16.76$ & $0.375$ & $44.24$\\
         \cmidrule{2-6}
         & Ours & $0.350$ & $\mathbf{14.26}$ & $0.367$ & $\mathbf{29.37}$\\
        \bottomrule
    \end{tabular}}
    \label{tab:kitti_zeroshot}
    \vspace{-10pt}
\end{table}

\noindent{}\textbf{KITTI~\cite{Geiger2012} benchmark.} \Cref{tab:kitti_benchmark} clearly shows the compelling performance of \ourmodel on the official KITTI private test set. We show the results of the latest published methods only. The table is from the official KITTI leaderboard.
\begin{table}[]
    \centering
    \caption{\textbf{Results on official KITTI~\cite{Geiger2012} Benchmark.} Comparison of performance of methods trained on KITTI and tested on the official KITTI private test set.}
    \vspace{-10pt}
    \resizebox{\columnwidth}{!}{
    \begin{tabular}{l|cccc}
    \toprule
    \multirow{2}{*}{\textbf{Method}} & $\mathbf{SI_{\log}}$ & $\mathbf{Sq.Rel}$ & $\mathbf{A.Rel}$ & $\mathbf{iRMS}$\\
     & \multicolumn{4}{c}{\textit{Lower is better}}\\
    \toprule
    PAP \cite{Zhang2019} & 13.08 & 2.72 \% & 10.27 \% & 13.95\\
    P3Depth \cite{Patil2022} & 12.82 & 2.53 \% & 9.92 \% & 13.71\\
    VNL \cite{Yin2019} & 12.65 & 2.46 \% & 10.15 \% & 13.02\\
    DORN \cite{Yu2019} & 11.77 & 2.23 \% & 8.78 \% & 12.98\\
    BTS \cite{Lee2019} & 11.67 & 2.21 \% & 9.04 \% & 12.23\\
    PWA \cite{Lee2021} & 11.45 & 2.30 \% & 9.05 \% & 12.32\\
    ViP-DeepLab \cite{Qiao2021} & 10.80 & 2.19 \% & 8.94 \% & 11.77\\
    NeWCRF \cite{Yuan2022} & 10.39 & 1.83 \% & 8.37 \% & 11.03\\
    PixelFormer \cite{Agarwal2022AttentionAE} & 10.28 & 1.82 \% & 8.16 \% & 10.84\\
    \midrule
    \midrule
    Ours (iDisc) & \textbf{9.89} & \textbf{1.77 \%} & \textbf{8.11 \%} & \textbf{10.73}\\
    \bottomrule
    \end{tabular}}
    \label{tab:kitti_benchmark}
    \vspace{-10pt}
\end{table}

\begin{table}[tb]
    \centering
    \caption{\textbf{Comparison on NYU with 3D metrics.} F1-score for varying threshold (m) and Chamfer distance (m) on point clouds.}
    \vspace{-10pt}
    \resizebox{\columnwidth}{!}{
    \begin{tabular}{lccccccc}
        \toprule
        \textbf{Method} & $\mathbf{F1}_{0.05} \uparrow$ & $\mathbf{F1}_{0.1} \uparrow$ & $\mathbf{F1}_{0.2} \uparrow$ & $\mathbf{F1}_{0.3} \uparrow$ & $\mathbf{F1}_{0.5} \uparrow$ & $\mathbf{F1}_{0.75} \uparrow$ & $\mathbf{D}_{\mathrm{Chamfer}} \downarrow$\\
        \toprule
        BTS~\cite{Lee2019} & $24.5$ & $47.0$ & $72.4$ & $84.4$ & $93.6$ & $97.2$ & $0.169$\\
        AdaBins~\cite{Bhat2020} & $24.0$ & $47.0$ & $73.0$ & $84.7$ & $94.0$ & $97.4$ & $0.163$\\
        NeWCRF~\cite{Yuan2022} & $25.5$ & $48.6$ & $74.0$ & $85.4$ & $94.4$ & $97.6$ & $0.156$\\
        \midrule
        iDisc & $\mathbf{27.8}$ & $\mathbf{52.0}$ & $\mathbf{77.0}$ & $\mathbf{87.8}$ & $\mathbf{95.5}$ & $\mathbf{98.1}$ & $\mathbf{0.131}$\\
        \bottomrule
    \end{tabular}}
    \label{tab:f1-score}
    \vspace{-10pt}
\end{table}

\begin{figure}[]
    \centering
    \includegraphics[width=\columnwidth]{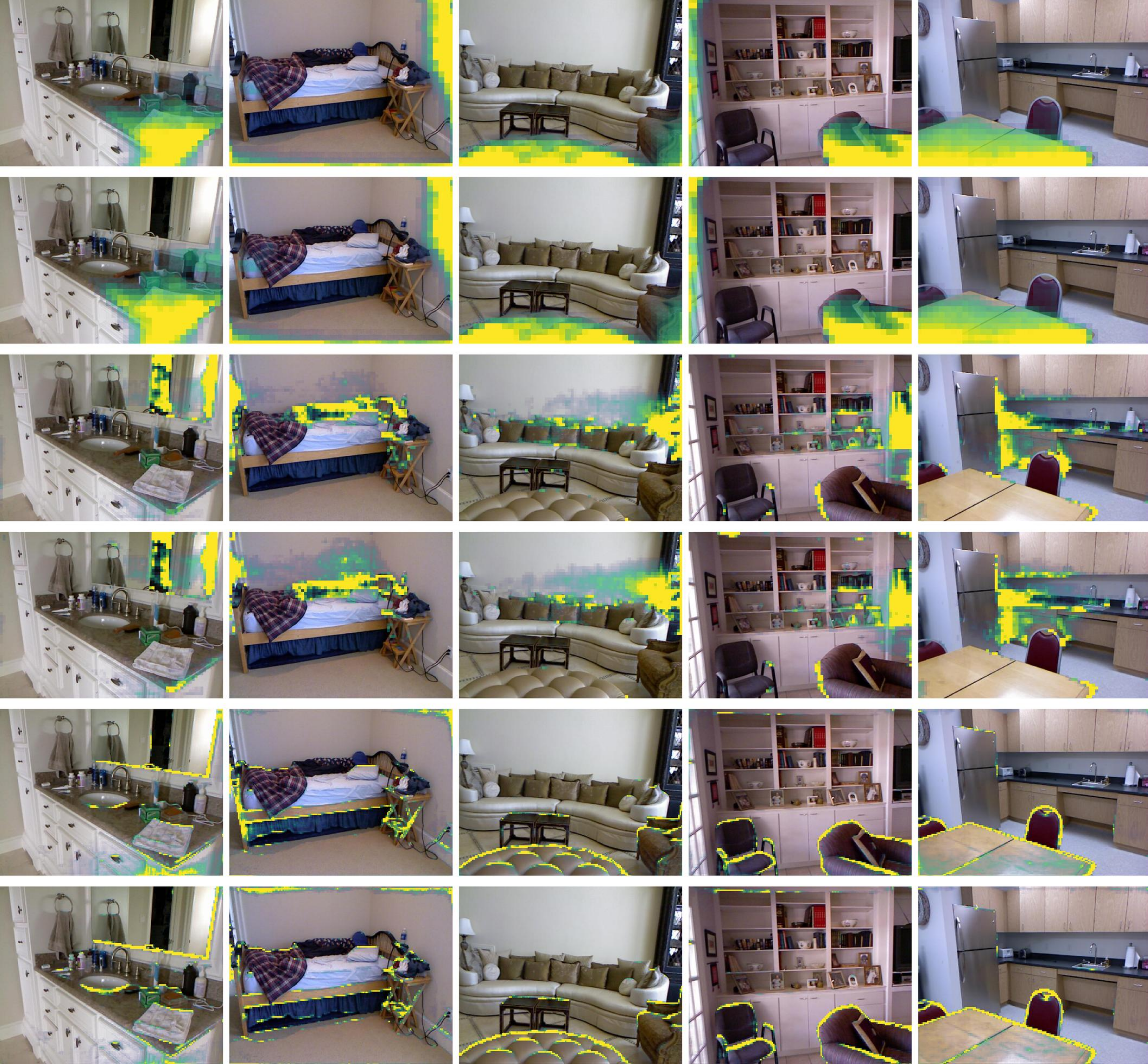}
    \vspace{-20pt}
    \caption{\textbf{Examples of attention maps degeneration.} Each pair of rows shows two different IDRs' attention maps, each pair is extracted from a different resolution. Some IDRs degenerate onto other IDRs, avoiding over-partitioning when more IDRs than those needed are utilized to represent the scene.}
    \label{fig:attention_collapse}
    \vspace{-10pt}
\end{figure}

\noindent{}\textbf{IDRs collapse.} We argue that our model is able to avoid over-clustering when performing the adaptive partitioning in AFP step. Over-clustering is the phenomenon occurring when the number of partitions enforced is more than the underlying true one. The \ourmodule module is able to avoid over-clustering by degenerating some IDRs onto others, thus not introducing any detrimental partition of the feature space. Degeneration of the same IDR is visible in \cref{fig:attention_collapse}. 

\begin{figure}[]
    % \vspace{-10pt}
    \centering
    \begin{subfigure}[b]{0.24\columnwidth}
        \centering
        \includegraphics[width=1\linewidth]{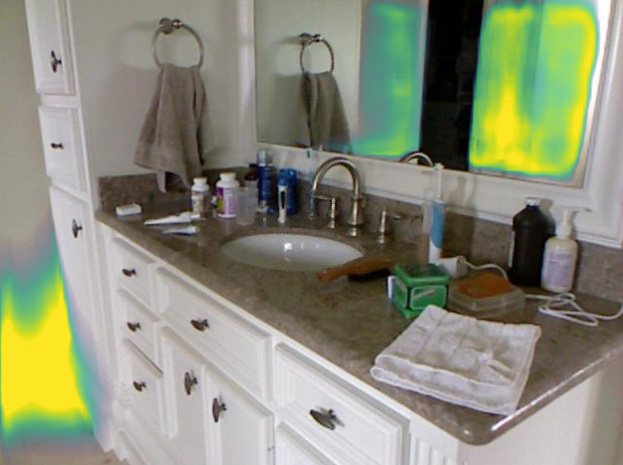}
    \end{subfigure}
    \begin{subfigure}[b]{0.24\columnwidth}
        \centering
        \includegraphics[width=1\linewidth]{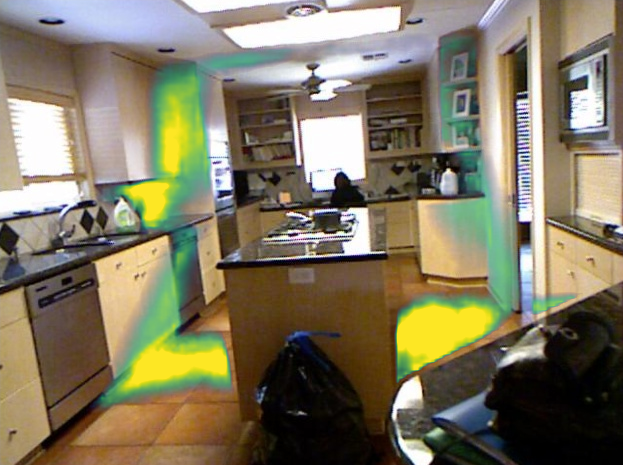}
    \end{subfigure}
    \begin{subfigure}[b]{0.24\columnwidth}
        \centering
        \includegraphics[width=1\linewidth]{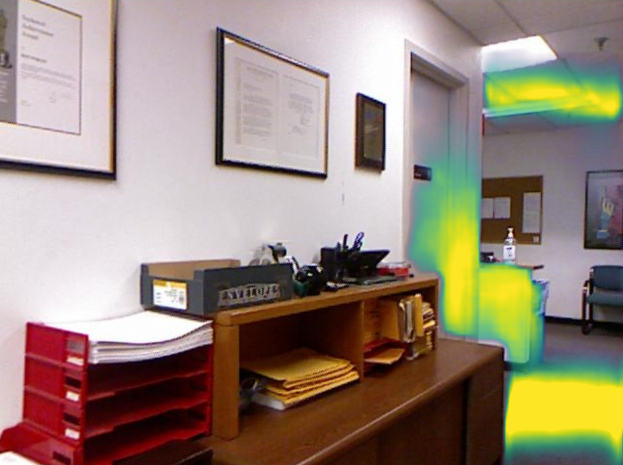}
    \end{subfigure}
    \begin{subfigure}[b]{0.24\columnwidth}
        \centering
        \includegraphics[width=1\linewidth]{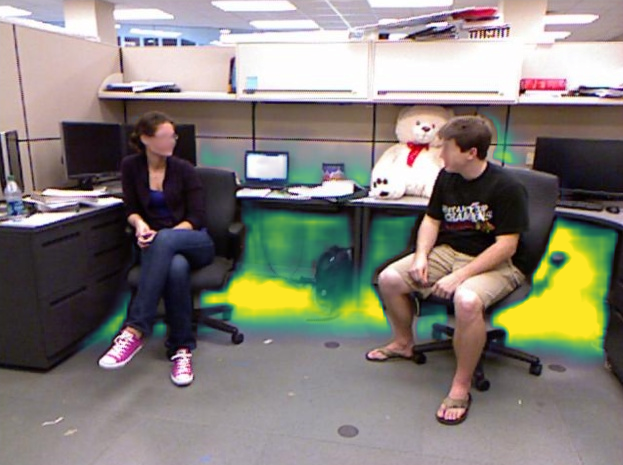}
    \end{subfigure}
    \par
    \begin{subfigure}[b]{0.24\columnwidth}
        \centering
        \includegraphics[width=1\linewidth]{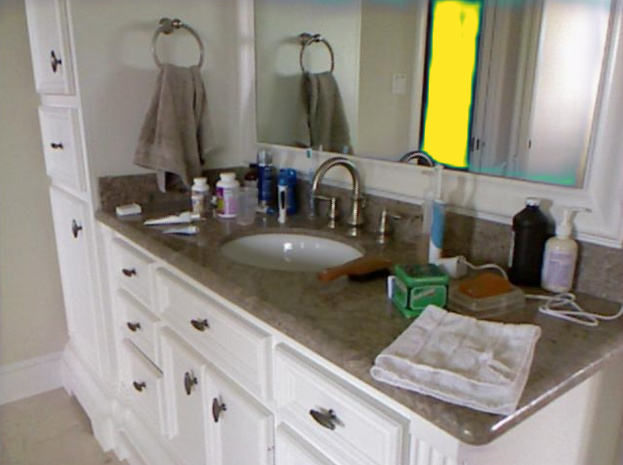}
    \end{subfigure}
    \begin{subfigure}[b]{0.24\columnwidth}
        \centering
        \includegraphics[width=1\linewidth]{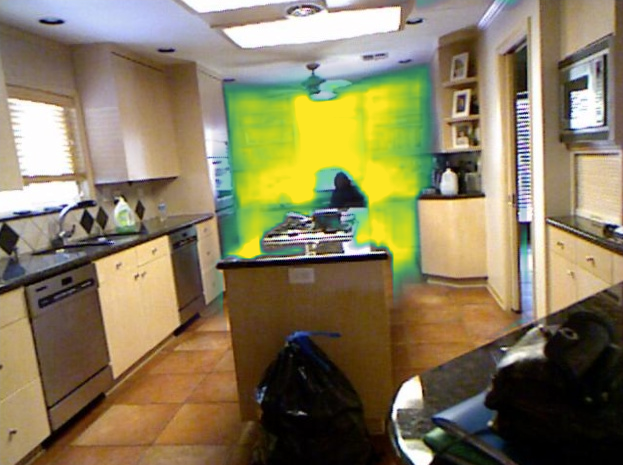}
    \end{subfigure}
    \begin{subfigure}[b]{0.24\columnwidth}
        \centering
        \includegraphics[width=1\linewidth]{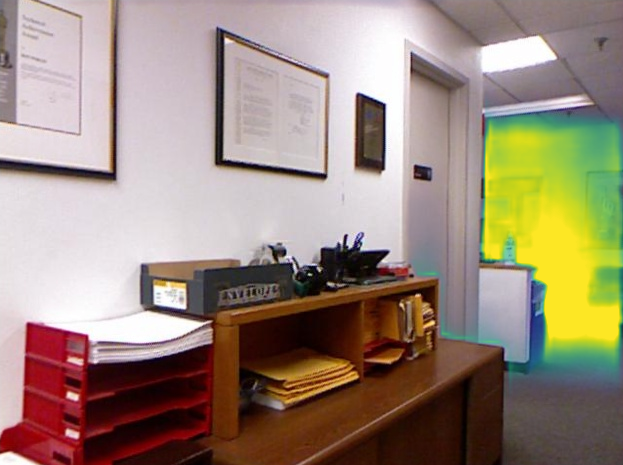}
    \end{subfigure}
    \begin{subfigure}[b]{0.24\columnwidth}
        \centering
        \includegraphics[width=1\linewidth]{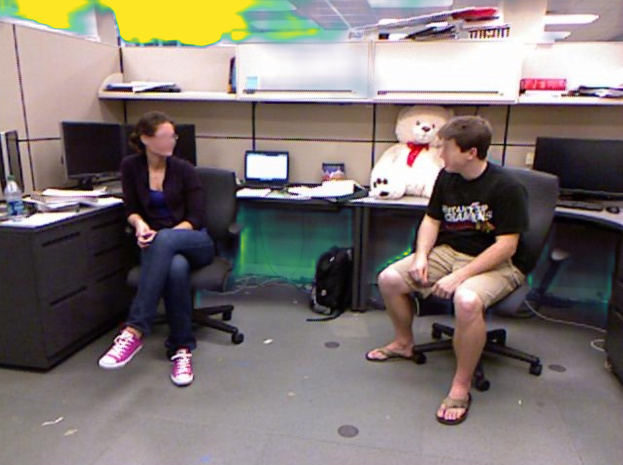}
    \end{subfigure}
    \par
    \begin{subfigure}[b]{0.24\columnwidth}
        \centering
        \includegraphics[width=1\linewidth]{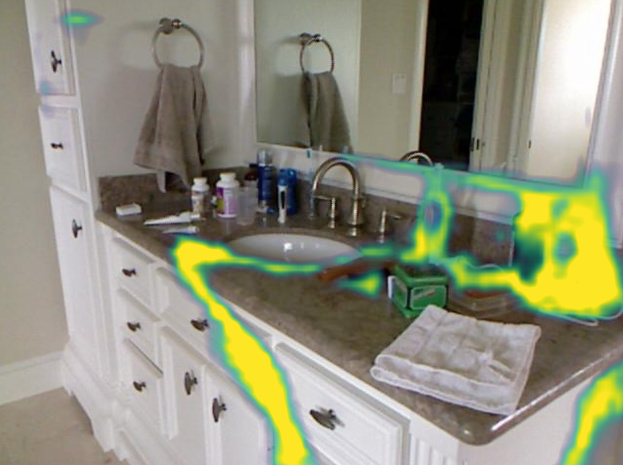}
    \end{subfigure}
    \begin{subfigure}[b]{0.24\columnwidth}
        \centering
        \includegraphics[width=1\linewidth]{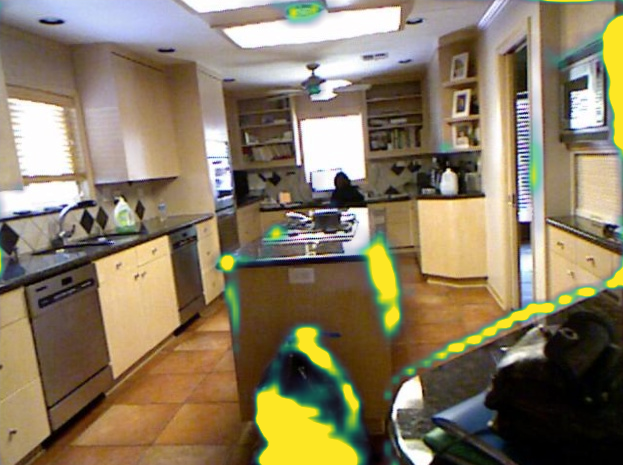}
    \end{subfigure}
    \begin{subfigure}[b]{0.24\columnwidth}
        \centering
        \includegraphics[width=1\linewidth]{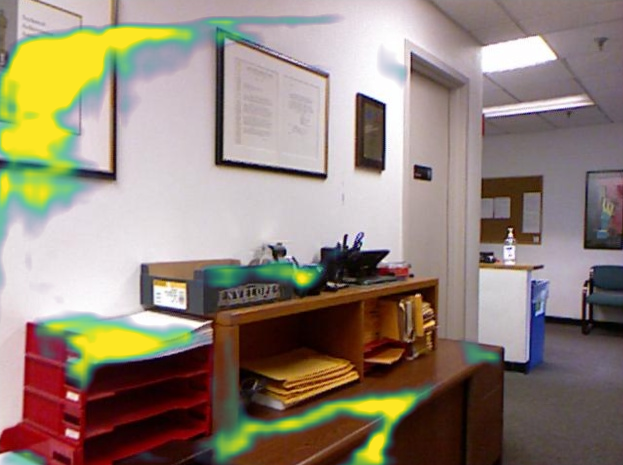}
    \end{subfigure}
    \begin{subfigure}[b]{0.24\columnwidth}
        \centering
        \includegraphics[width=1\linewidth]{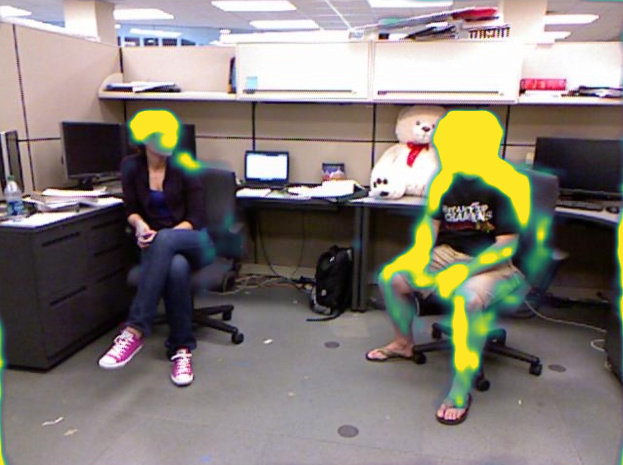}
    \end{subfigure}\\
    \vspace{-10pt}
    \caption{\textbf{Attention visualization.} Attention maps of three different IDRs at mid-resolution, on four different images from NYU.}
    \label{fig:vis}
\vspace{-10pt}
\end{figure}

\noindent{}\textbf{Attention depth planes.} \cref{fig:vis} shows three IDRs (each row shows a specific IDR, as in main paper figures) at the middle resolution. The top two rows support the ``speculation'' on iDisc's ability to still capture depth planes.

\noindent{}\textbf{Computational complexity.} We provide the analysis of the components in~\Cref{tab:resources}. Removing MSDA increases throughput to $20$fps, with only a slight loss in performance.
% (cf.~\Cref{tab:ablations}). 
Note that our implementation is not fully optimized for performance. NeWCRF~\cite{Yuan2022} uses the same backbone but more parameters and similar throughput to iDisc without MSDA.
\begin{table}[]
    % \vspace{-10pt}
    \centering
    \caption{\textbf{Computational complexity analysis} on an RTX 3090 with input images of size 640$\times$480 and SWin-L backbone.}
    \vspace{-10pt}
    \resizebox{\columnwidth}{!}{
    \footnotesize
    \begin{tabular}{lccc}
        \toprule
        \textbf{Component} & \textbf{Latency} (ms) & \textbf{Throughput} (fps) & \textbf{Parameters} (M)\\
        \toprule
        Encoder & $23.6$ & $42.4$ & $194.9$\\
        MSDA & $72.8$ & $13.7$ & $2.83$\\
        FPN & $2.7$ & $370.5$ & $4.11$\\
        AFP & $12.4$ & $80.7$ & $2.78$\\
        ISD & $9.6$ & $103.7$ & $4.59$\\
        \midrule
        iDisc (w/o MSDA) & $48.2$ & $20.7$ & $206.4$\\
        iDisc & $121.1$ & $8.3$ & $209.2$\\
        \bottomrule
    \end{tabular}}
    \label{tab:resources}
    \vspace{-10pt}
\end{table}

\section{Ablations}

\begin{figure}[h]
    \centering
    \includegraphics[width=\columnwidth]{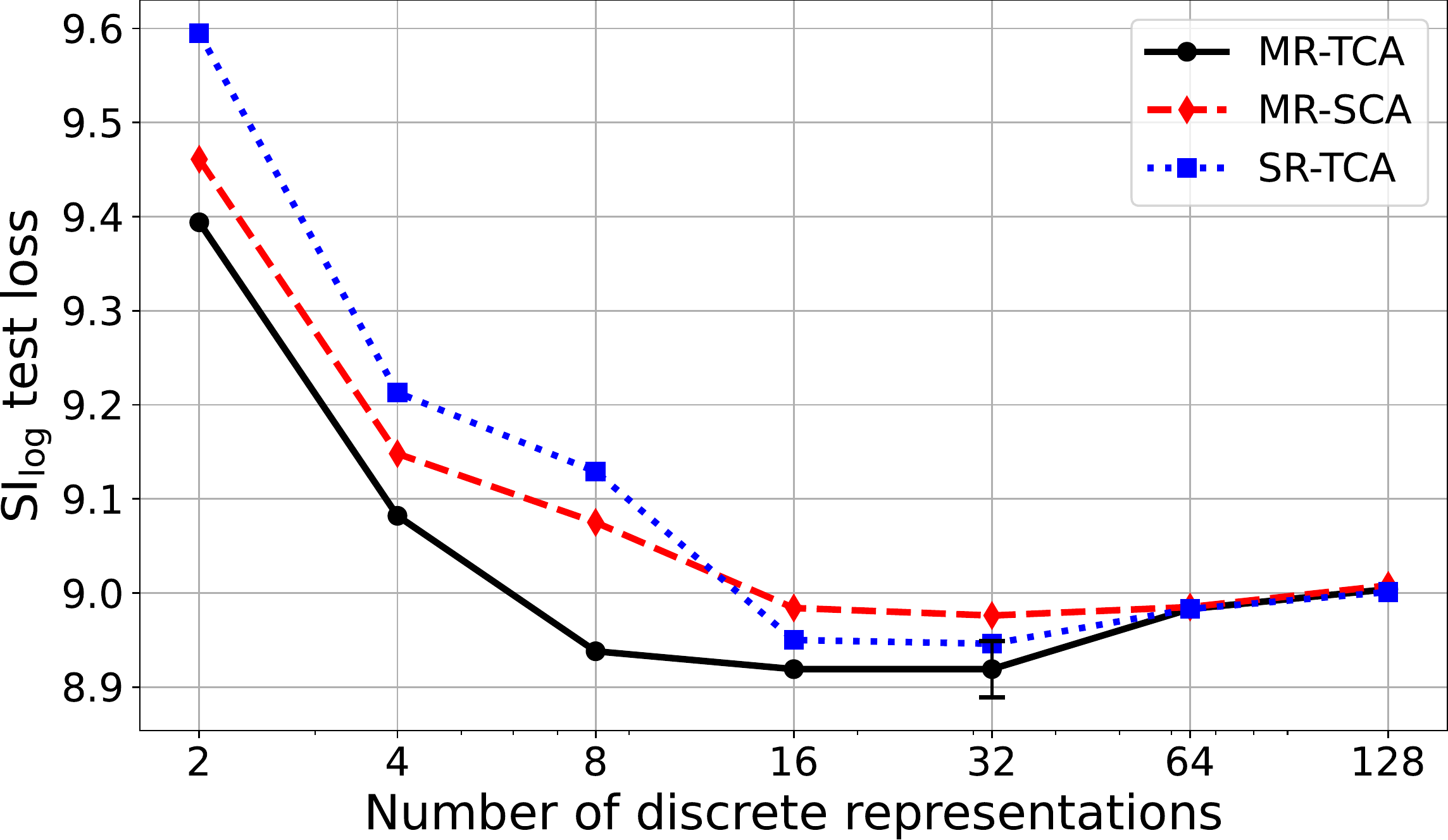}
    \vspace{-10pt}
    \caption{\textbf{Ablations on the number of IDRs and \ourmodule module's configurations.} MR-TCA: Multi-Resolution and Transposed cross-attention, MR-SCA: Multi-Resolution and Standard cross-attention in AFP, Single-Resolution and Transposed cross-attention. MR-TCA corresponds to the \ourmodel model. MR-SCA corresponds to using cross-attention instead of cluster-inducing transposed attention. SR-TCA corresponds to having only one intermediate representation, namely the final depth directly. The error bar in correspondence of 32 on the x-axis indicates the standard deviation.}
    \label{fig:ablations_tot}
    \vspace{-10pt}
\end{figure}

\noindent{}\textbf{Number of IDRs.} We ablate the model with respect to the number of IDRs exploited by \ourmodel. In particular,  we sweep the number of IDRs between 2 and 128 with a base-two log scale. The black-solid line in \cref{fig:ablations_tot} shows the trend of \ourmodel when ablating the IDRs: the optimum is reached in the interval [8, 32]. When more representations are utilized, we argue that noise is introduced in the bottleneck and the discretization process is not actually enforced. The discretization does not occur since the number of IDRs would be close to the number of feature map elements. On the other hand, 2 or 4 IDRs are already enough to obtain decent results, although not particularly visually appealing. In particular, we speculate that the extreme case of utilizing two IDRs can lead to the model representing the maximum depth with one of the two representations and the minimum one with the other. Therefore, the model is still able to interpolate between the depth interval range. The interpolation occurs thanks to the convex combination, defined by $\mathrm{softmax}$, of maximum and minimum depth. More specifically, $\mathrm{softmax}$ is guided by the similarity between the pixel embeddings and the corresponding depth representations. Thus, the model is virtually able to define the full depth range via the weights of the $\mathrm{softmax}$ convex combination modulated by the pixel embeddings. When utilizing only one representation, the model does not converge, if not to the mean scene depth.

\noindent{}\textbf{Single resolution in ISD.} The dotted-blue line in \cref{fig:ablations_tot} shows the trend when only one resolution is processed in the ISD stage of the \ourmodule module. In such a configuration, the output of the \ourmodule module is directly the depth. Here, no fusion is to be performed between different intermediate representations. One can observe that single-resolution is particularly affected when few IDRs are utilized. We argue that multi-resolution counterparts can compensate for the diminished granularity of internal representation. The compensation stems from combining different facets, \ie, at different resolutions, of the IDRs.

\noindent{}\textbf{Attention in AFP.} The dashed-red line in \cref{fig:ablations_tot} shows the performance when standard cross-attention is utilized in AFP, instead of the partition-inducing transposed cross-attention. In this case, a high number of IDRs does not affect performance. Here, the IDRs are additive instead of soft mutually exclusive, \ie, the IDRs from transposed cross-attention. Therefore, utilizing more IDRs is virtually not detrimental.

\noindent{}\textbf{\ourmodule module layers and iterations.} \Cref{tab:ablations_iters} shows the ablation study on the iterations and layers utilized in the stages of the \ourmodule module. We can observe that a higher number of transposed cross-attention, thus of iterative partitioning, has almost no effect on performances, since the partitions have probably converged. On the other hand, when $\mathrm{N_{AFP}}$ is one, results are similar to using only the IDRs priors since the adaptive part is truncated too early. Iterations of ISD stage ($\mathrm{N_{ISD}}$) correspond to the number of cross-attention layers utilized in the last stage of the \ourmodule module. \ourmodel is already able to obtain good results with only one layer, while increasing the layers may lead to overfitting. Nonetheless, \Cref{tab:afp-abl_zeroshot} clearly shows how the input-dependency in the feature partitioning, \ie, $\mathrm{N_{AFP}}$ greater than zero, leads to improved generalization.

\begin{table}[]
    \centering
    \caption{\textbf{Ablations of \ourmodule module iterations.} $\mathrm{N_{AFP}}$: number of iterations in the AFP stage, $\mathrm{N_{ISD}}$: number of cross-attention layers in ISD stage. The last row corresponds to the architecture utilized for all other experiments.}
    \vspace{-10pt}
    \resizebox{\columnwidth}{!}{
    \begin{tabular}{ccc|ccc}
        \toprule
        & $\mathrm{N_{AFP}}$ & $\mathrm{N_{ISD}}$ & $\boldsymbol{\delta_1} \uparrow$ & $\mathbf{RMS} \downarrow$ & $\mathbf{A.Rel} \downarrow$\\
        \toprule    
        1 & 2 & 1 & $0.938$ & $0.314$ & $0.086$\\
        2 & 2 & 3 & $0.934$ & $0.316$ & $0.088$\\
        3 & 2 & 4 & $0.935$ & $0.317$ & $0.089$\\
        \midrule
        4 & 1 & 2 & $0.935$ & $0.317$ & $0.087$\\
        5 & 3 & 2 & $0.938$ & $0.313$ & $0.086$\\
        6 & 4 & 2 & $0.938$ & $0.314$ & $0.086$\\
        \midrule
        7 & 2 & 2 & $0.940$ & $0.313$ & $0.086$\\
        \bottomrule
    \end{tabular}
    }
    \label{tab:ablations_iters}
    \vspace{-5pt}
\end{table}

\begin{table}[]
    % \vspace{-10pt}
    \centering
    \caption{\textbf{Test loss for varying $\mathrm{N_{AFP}}$.} The models are trained on NYU and tested on the ``Test Dataset''.}
    \vspace{-10pt}
    \resizebox{\columnwidth}{!}{
    \footnotesize
    \begin{tabular}{lccc}
        \toprule
        \textbf{Test Dataset} & $\mathrm{SI_{\log}}@\mathrm{N_{AFP}}=0$ & $\mathrm{SI_{\log}}@\mathrm{N_{AFP}}=1$ & $\mathrm{SI_{\log}}@\mathrm{N_{AFP}}=2$\\
        \toprule
        NYU & $10.43$ & $9.471$ & $8.845$\\
        SUN-RGBD & $12.76$ & $11.50$ & $10.91$\\
        Diode & $20.97$ & $18.97$ & $18.11$\\
        \bottomrule
    \end{tabular}}
    \label{tab:afp-abl_zeroshot}
    \vspace{-10pt}
\end{table}

\section{Network Architecture}
\noindent{}\textbf{Encoder.} We show the effectiveness of our method with different encoders, both convolutional and transformer-based ones, \eg, ResNet~\cite{He2015}, EfficientNet~\cite{Tan2019} and SWin~\cite{Liu2021}. However, all of them follow the same structure, where the receptive field of either convolution or windowed attention is increased by decreasing the resolution of the feature maps. The final size of the feature map is 1/32 of the input image. All backbones utilized are originally designed for classification, thus we remove the last 3 layers, \ie, the pooling layer, fully connected layer, and $\mathrm{softmax}$ layer. We employ each backbone to generate feature maps of different resolutions, which can be used as skip connections to the decoder.

\noindent{}\textbf{Multi-scale deformable attention refinement.} The feature maps at different resolutions are refined via mutli-scale deformable attention~\cite{Zhu2021DefDETR}. Deformable attention efficiency relies on attending only a few locations to compute attention for each pixel, instead of having full connectivity likewise standard attention. Deformable attention is also utilized to share information at different resolutions. Each layer is composed of layer normalization~\cite{Ba2016} (LN), fully connected layers (FC), and Gaussian Error Linear Unit~\cite{Hendrycks2016} (GeLU).

\noindent{}\textbf{Decoder.} Feature maps at different resolutions are combined via a feature pyramidal network (FPN) which exploits LN, GeLU activations, and convolutional layers with  3$\times$3 kernels. The decoder outputs at different resolutions correspond to the set of pixel embeddings ($\mathcal{P}$). 

\noindent{}\textbf{AFP and ISD.} AFP stage is an iterative component, thus weights are shared across layers. One layer comprises transposed cross-attention, LN, GeLU activations, and FC layers: three dedicated layers for key, queries and value tensors, and one layer applied to the attention layer output. The architectural components of the ISD stage are the same as AFP's components, except for the use of standard cross-attention instead of transposed one, and the weights are not shared.

\section{Visualizations}
\begin{figure}[h]
    \centering
    \includegraphics[width=\columnwidth]{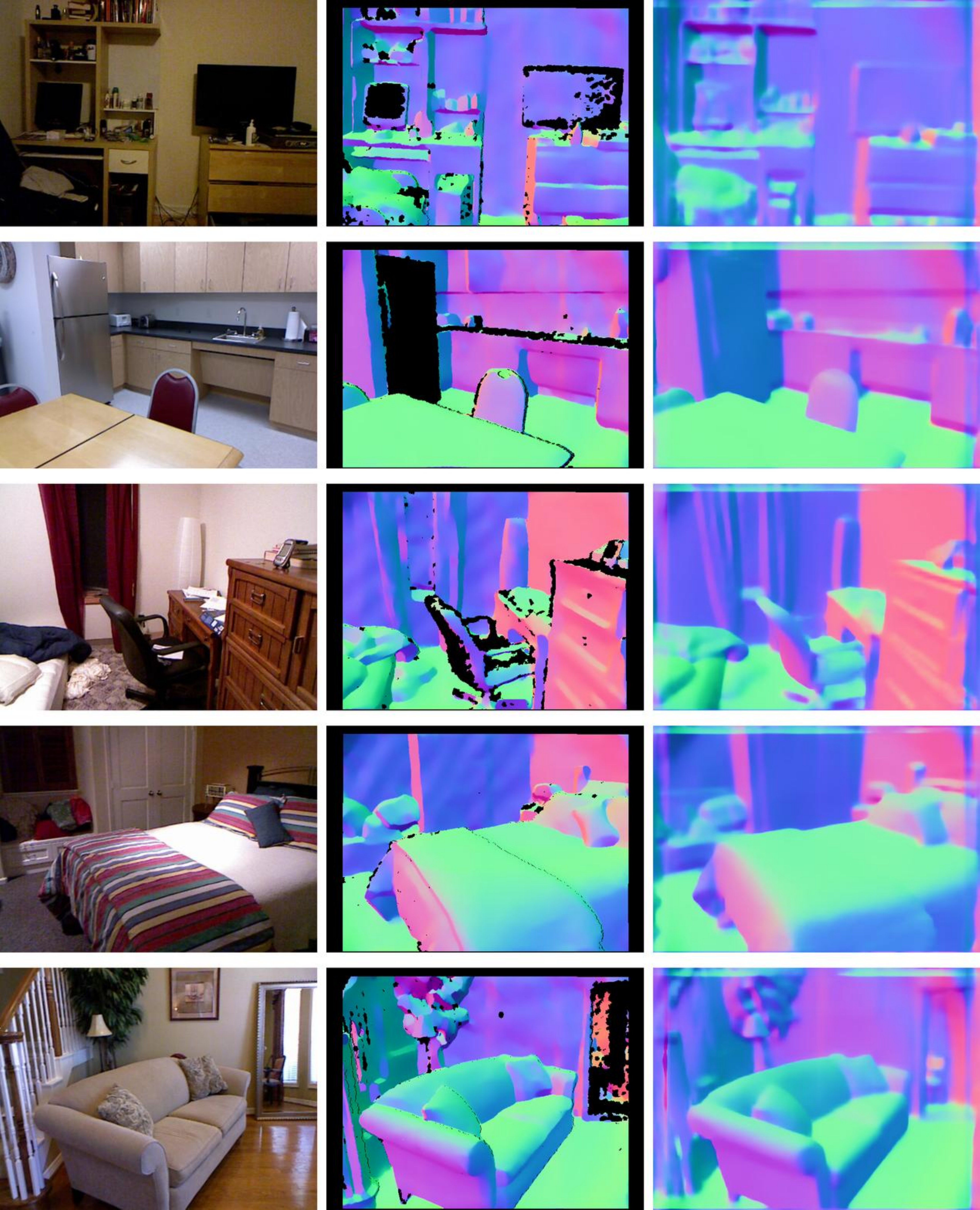}\\
    \footnotesize
    \begin{tabularx}{\columnwidth}{l}
        \hspace{0.1\columnwidth}\textbf{Image}\hspace{0.26\columnwidth}\textbf{GT}\hspace{0.28\columnwidth}\textbf{Ours}
    \end{tabularx}\\
    \vspace{-10pt}
    \caption{\textbf{Qualitative results on NYU for surface normals estimation.} Each row corresponds to one test sample from NYU. The first two columns correspond to the input image and depth GT, respectively. The third column is the predicted normals of the tangent plane for every pixel.}
    \vspace{-15pt}
\end{figure}
\begin{figure*}
    \centering
    \includegraphics[width=\textwidth]{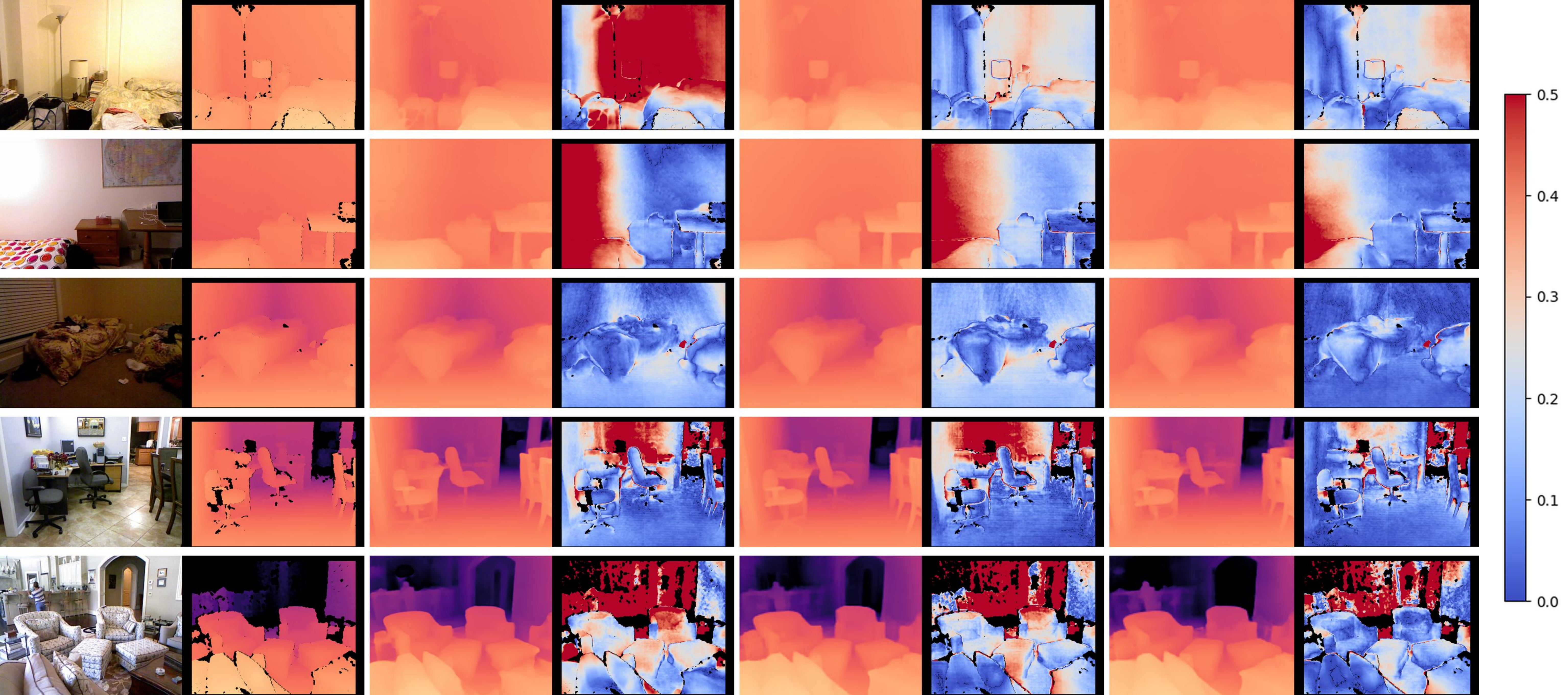}\\
    \footnotesize
    \begin{tabularx}{\textwidth}{l}
        \hspace{0.025\textwidth}\textbf{Image}\hspace{0.085\textwidth}\textbf{GT}\hspace{0.13\textwidth}\textbf{AdaBins}\cite{Bhat2020}\hspace{0.15\textwidth}\textbf{NeWCRF}\cite{Yuan2022}\hspace{0.16\textwidth}\textbf{Ours}
    \end{tabularx}\\
    \vspace{-5pt}
    \caption{\textbf{Qualitative results on NYU.} Each row corresponds to one test sample from NYU. The first two columns correspond to the input image and depth GT, respectively. Each couple afterward corresponds to the pair output depth and error map. Error maps are clipped at 0.5m and the corresponding colormap is \textit{coolwarm}.}
\end{figure*}

\begin{figure*}[]
    \centering
    \includegraphics[width=\textwidth]{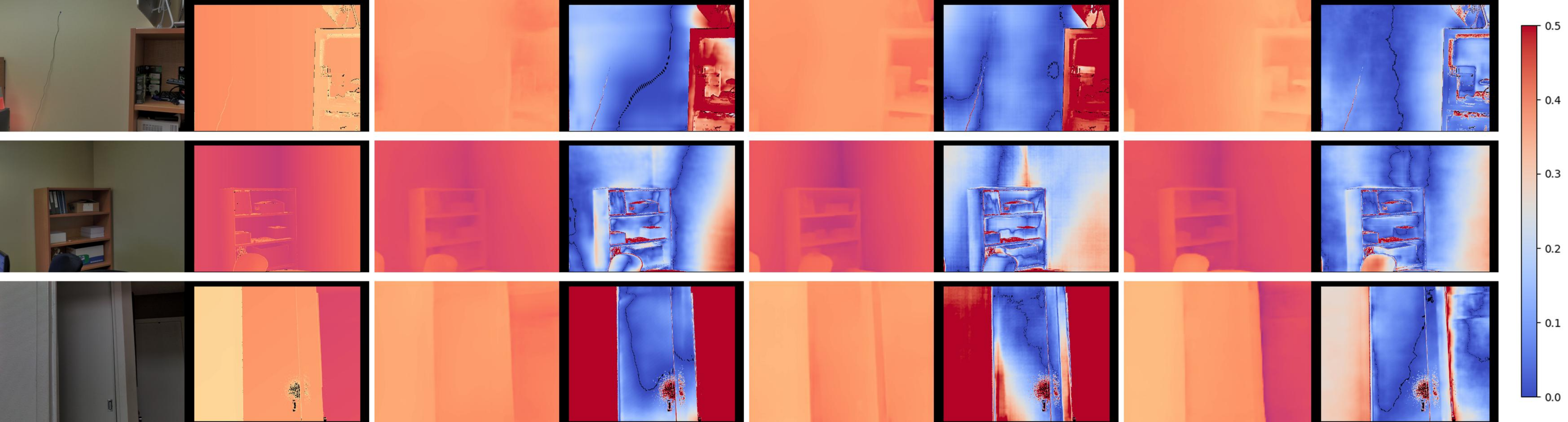}\\
    \footnotesize
    \begin{tabularx}{\textwidth}{l}
        \hspace{0.025\textwidth}\textbf{Image}\hspace{0.085\textwidth}\textbf{GT}\hspace{0.13\textwidth}\textbf{AdaBins}\cite{Bhat2020}\hspace{0.15\textwidth}\textbf{NeWCRF}\cite{Yuan2022}\hspace{0.16\textwidth}\textbf{Ours}
    \end{tabularx}\\
    \vspace{-5pt}
    \caption{\textbf{Qualitative results on Diode.} Each row corresponds to one zero-shot test sample for the model trained on NYU and tested on Diode. The first two columns correspond to the input image and depth GT, respectively. Each subsequent couple corresponds to the pair output depth and error map. Error maps are clipped at 0.5m and the corresponding colormap is \textit{coolwarm}.}
\end{figure*}

\begin{figure*}[h]
    \centering
    \includegraphics[width=\textwidth]{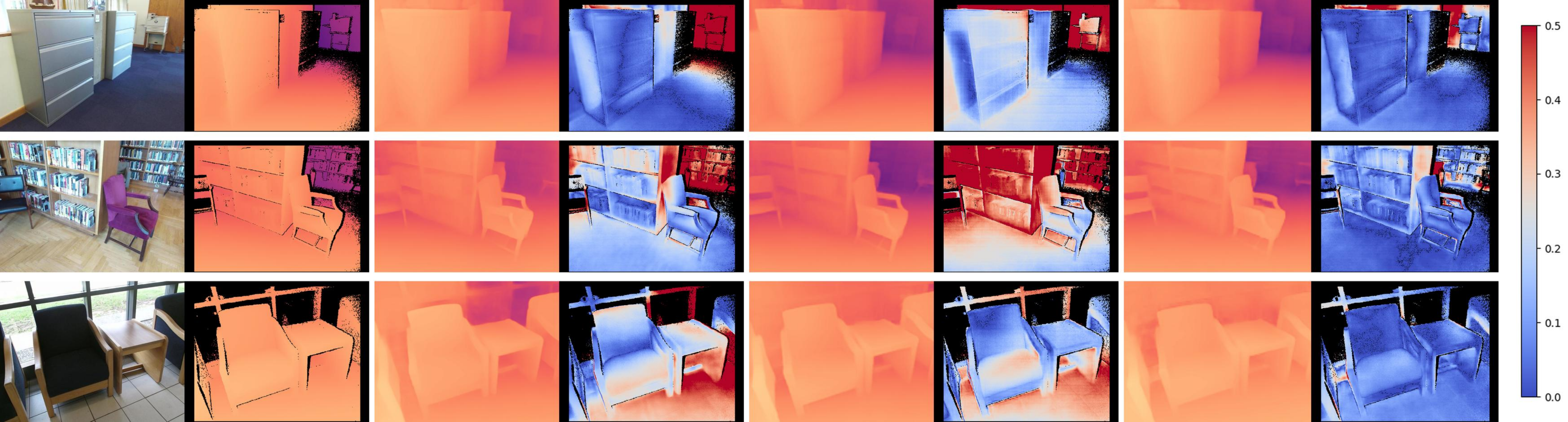}\\
    \footnotesize
    \begin{tabularx}{\textwidth}{l}
        \hspace{0.025\textwidth}\textbf{Image}\hspace{0.085\textwidth}\textbf{GT}\hspace{0.13\textwidth}\textbf{AdaBins}\cite{Bhat2020}\hspace{0.15\textwidth}\textbf{NeWCRF}\cite{Yuan2022}\hspace{0.16\textwidth}\textbf{Ours}
    \end{tabularx}\\
    \vspace{-5pt}
    \caption{\textbf{Qualitative results on SUN-RGBD.} Each row corresponds to one zero-shot test sample for the model trained on NYU and tested on SUN-RGBD. The first two columns correspond to the input image and depth GT, respectively. Each subsequent couple corresponds to the pair output depth and error map. Error maps are clipped at 0.5m and the corresponding colormap is \textit{coolwarm}.}
\end{figure*}

\begin{figure*}[h]
    \centering
    \includegraphics[width=\textwidth]{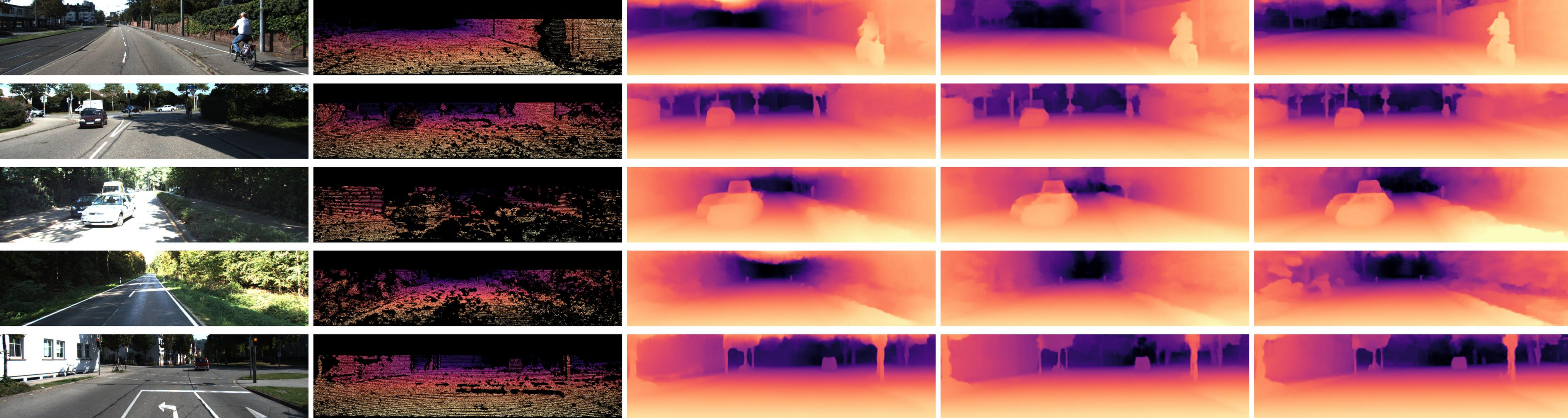}\\
    \footnotesize
    \begin{tabularx}{\textwidth}{l}
        \hspace{0.065\textwidth}\textbf{Image}\hspace{0.17\textwidth}\textbf{GT}\hspace{0.15\textwidth}\textbf{AdaBins}\cite{Bhat2020}\hspace{0.11\textwidth}\textbf{NeWCRF}\cite{Yuan2022}\hspace{0.14\textwidth}\textbf{Ours}
    \end{tabularx}\\
    \vspace{-5pt}
    \caption{\textbf{Qualitative results on KITTI.} Each row corresponds to a test sample from KITTI. The first two columns correspond to the input image and depth GT, respectively. The following columns correspond to the respective models trained on KITTI.}
\end{figure*}

\begin{figure*}[h]
    \centering
    \includegraphics[width=\textwidth]{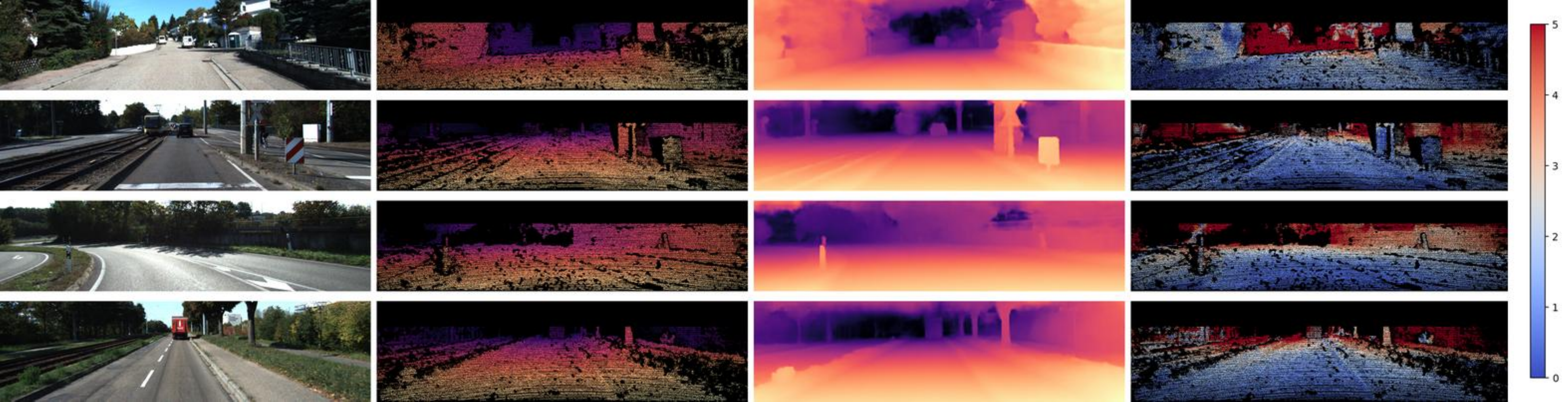}\\
    \footnotesize
    \begin{tabularx}{\textwidth}{l}
        \hspace{0.08\textwidth}\textbf{Image}\hspace{0.21\textwidth}\textbf{GT}\hspace{0.212\textwidth}\textbf{Ours}\hspace{0.21\textwidth}\textbf{Error}
    \end{tabularx}\\
    \vspace{-5pt}
    \caption{\textbf{Failure cases on KITTI.} Each row corresponds to one test sample from KITTI Eigen-split validation set. The examples selected correspond to the four worst samples in terms of absolute error. Error maps are clipped at 5m and the corresponding colormap is \textit{coolwarm}.}
\end{figure*}

\begin{figure*}[h]
    \centering
    \includegraphics[width=\textwidth]{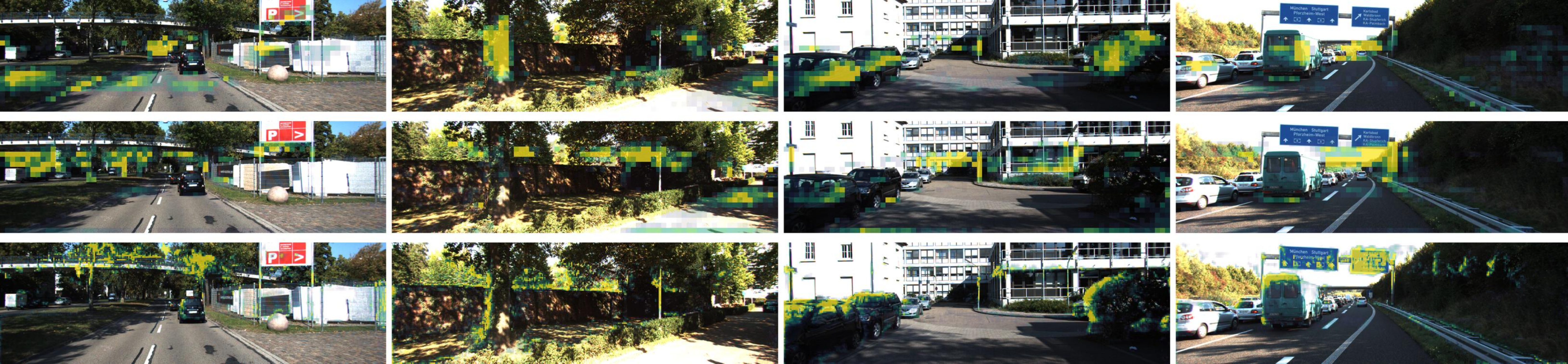}
    \vspace{-5pt}
    \caption{\textbf{Attention maps on KITTI for three different IDRs.} Each row presents the attention map of a specific IDR for four test images. Each IDR focuses on a specific high-level concept. The first two rows pertain to IDR at the lowest resolution while the last corresponds to the highest resolution. Best viewed on a screen and zoomed in.}
\end{figure*}

